\documentclass{article}

\usepackage{spconf}
\usepackage[cmex10]{amsmath}
\usepackage{amsthm}
\usepackage{amssymb}
\usepackage{mathrsfs}
\usepackage{graphicx}
\usepackage{float}
\usepackage{array}
\usepackage{epstopdf}
\usepackage{multirow}

\usepackage{blkarray}
\usepackage{mathtools}
\usepackage{times}
\usepackage{subcaption}
\usepackage[T1]{fontenc}

\newcommand{\argmin}{\arg\!\min}

\title{Screen Content Image Segmentation Using Least Absolute Deviation Fitting}
\name{Shervin Minaee and Yao Wang}
\address{Department of Electrical and Computer Engineering, Polytechnic School of Engineering, \\ New York University, NY, USA.}
\begin{document}
%\ninept
%
\maketitle
\begin{abstract}
We propose an algorithm for separating the foreground (mainly text and line graphics) from the smoothly varying background in screen content images. The proposed method is designed based on the assumption that the background part of the image is smoothly varying and can be represented by a linear combination of a few smoothly varying basis functions, while the foreground text and graphics create sharp discontinuity and cannot be modeled by this smooth representation. The algorithm separates the background and foreground using a least absolute deviation method to fit the smooth model to the image pixels. This algorithm has been tested on several images from HEVC standard test sequences for screen content coding, and is shown to have superior performance over other popular methods, such as k-means clustering based segmentation in DjVu and shape primitive extraction and coding (SPEC) algorithm. Such background/foreground segmentation are important pre-processing steps for text extraction and separate coding of background and foreground for compression of screen content images.
\end{abstract}

\section{Introduction}
\label{sec:intro}

%Background/foreground separation has several applications in image processing and video compression. By decomposing an image into background and foreground we will derive two layers which are easier to be processed and coded. Background layer usually contains the smooth part of the image and the foreground contains text/graphics and lines. This technique can be used for screen content video coding \cite{SCC}, text extraction \cite{text}, medical image segmentation and classification \cite{chromosome}, \cite{hojat} and principal line extraction from palmprint images \cite{palmprint}.

Screen content images refer to images appearing on the display screens of electronic devices. These images share similar characteristics as mixed content documents (such as a magazine page). They often contain two layers, pictorial background and the foreground consisting of text and line graphics. The usual image compression algorithm such as JPEG2000 \cite{jpeg} and HEVC intra frame coding \cite{HEVC} may not result in a good compression rate for this kind of images. 
%The reason being that, there are a lot of sharp edges in these images, therefore after taking the transform of these images, the energy cannot be compacted in a few coefficients and will be distributed in most of the transform domain coefficients, therefore results in a high bit-rate. 
In these cases, segmenting the image into two layers and coding them separately may be more efficient.
The idea of segmenting an image for better compression was proposed for check image compression \cite{check}, in DjVu algorithm for scanned document compression \cite{djvu} and the mixed raster content representation \cite{video_segmentation_coding}.

Screen content images are hard to segment, because the foreground may be overlaid over a smoothly varying background that has a color range that overlaps with the color of the foreground. Also because of the use of sub-pixel rendering, the same text/line often has different colors. Even in the absence of sub-pixel rendering, pixels belonging to the same text/line often has somewhat different colors.

Most of previous works regarding foreground segmentation are based on color clustering or color counting and have difficulty for cases where the background color has a large dynamic range or is similar to the foreground color in some regions. The hierarchical k-means clustering algorithm initially proposed in DjVu \cite{djvu} is a representative algorithm based on color clustering.
This algorithm applies the k-means clustering algorithm with k=2 on blocks in multi-resolution. It first applies the k-means clustering algorithm on a large block to obtain foreground and background colors and then uses them as the initial foreground and background colors for the smaller blocks in the next stages. This algorithm has difficulty for the regions where background and foreground color intensities overlap. 

In the shape primitive extraction and coding (SPEC) method, which was developed for segmentation of screen content \cite{SPEC}, a two-step segmentation algorithm is proposed. In the first step they classify each block of size $16 \times 16$ into either pictorial block or text/graphics based on the number of colors. If the number of colors is more than a threshold, it will be classified into pictorial block, otherwise to text/graphics.
In the second step, they refine the segmentation result of pictorial blocks, by extracting shape primitives (horizontal line, vertical line or a rectangle with the same color) and then comparing the size and color of the shape primitives with some threshold.
Because blocks containing smoothly varying background over a narrow range can also have a small color number,  it is hard to find a fixed color number threshold that can robustly separate pictorial blocks and text/graphics blocks. Furthermore, text and graphics  in screen content images typically have some variation in their colors, even in the absence of sub-pixel rendering. These  challenges limit the effectiveness of SPEC.
%This algorithm suffers from the fact that all blocks of size $16 \times 16$ with less than 32 different colors are classified into text/graphics which could also be true for a lot background blocks.

In this paper, we propose a foreground/background separation algorithm which uses the smoothness property of the background  and the fact that the foreground pixels typically deviate greatly from the smooth function fit to the background. Using this intuition we propose to use least absolute deviation approach \cite{LAD} to fit each image block using a smooth model.
Those pixels which can be represented with small distortion using this smooth model will be considered as background and the rest as foreground.
This technique can be used for screen content video coding \cite{SCC}, text extraction \cite{text}, medical image segmentation and classification \cite{chromosome}, \cite{hojat} and principal line extraction from palmprint images \cite{palmprint1}, \cite{palmprint2}.

%can decompose an image into two layers, background and foreground. Background contains the smooth part of the image and foreground contains any component with sharp edge and different structure from majority of the image, such as texts, graphics and lines. Least absolute deviations \cite{LAD} approach is used to fit a smooth model into the image. Those pixels which can be represented with small distortion using this smooth model will be considered as background and the rest as foreground. 
%The proposed segmentation algorithm can be used for different applications such as text extraction \cite{text}, principle line extraction in palmprint recognition and also image and video compression \cite{video_segmentation_coding}, medical image segmentation \cite{chromosome}. 
%The segmentation result for some images are shown in Figure 1.

\section{Least Absolute Deviation}
\label{SectionII}

In this paper, we look at the foreground segmentation problem from signal decomposition point of view. We assume that the background part of the image can be well represented with a simple smooth model, whereas the foreground pixels cannot be represented accurately with this smooth model. By well representation we mean that the distortion between the approximated smooth model and the actual pixel values is less than a desired threshold. To be more specific, we divide each image into non-overlapping blocks of size N$\times$N, and then represent each image block denoted by $F(x,y)$, with a smooth model $S(x,y;\alpha_1,...,\alpha_K)$, where $x$ and $y$ denote the horizontal and vertical axes and $\alpha_1,...,\alpha_K$ denote the parameters of this smooth model. 
For color images, $F(x,y)$ represents the luminance component.
In order to find the optimal model parameters, $\alpha_k$'s, we should define some cost function which measures the goodness of fit between the intensity of background pixels in the original image and the one predicted by smooth model, and then minimize the cost function as:
\begin{gather*}
\{\alpha_1^*,...,\alpha_K^*\}= \argmin_{\alpha_1,...,\alpha_K} \| F(x,y)- S(x,y;\alpha_1,...,\alpha_K)  \|_p
\end{gather*}
Now two questions should be answered:
\begin{enumerate}
\item What is an optimal smooth model for background layer representation.
\item What error measure (i.e. p value) to use such that the model parameters are mainly found using the background pixels.
\end{enumerate}
For the first question, we propose to use a linear combination of $K$ smooth basis functions $\sum_{k=1}^K \alpha_k P_k(x,y)$, where $P_k(x,y)$ denotes a 2D smooth basis function. 
%Since this model is a linear function of parameters, $\alpha_k$, it is simpler to find the optimal weights. 
Here we use a set of low frequency two-dimensional DCT basis functions, since they have been shown to be very efficient for image representation \cite{DCT}. The 2-D DCT function is defined as:
\begin{equation*}
P_{u,v}(x,y)= \beta_u \beta_v cos((2x+1)\pi u/2N) cos((2y+1)\pi u/2N) 
\end{equation*}
where $u$ and $v$ denote the frequency of the basis.
% and $x$ and $y$ denote spatial coordinate of the image pixel.
We order all the possible basis functions in the conventional zig-zag order in the (u,v) plane, and choose the first K basis functions.
%We collected several smooth background images, extracted all blocks of size $64\times64$ and tried to represent them with the first K DCT basis functions in zigzag order as $\sum_{k=1}^K \alpha_k P_k(x,y)$. 
%The reconstruction RMSEs (root MSE) as a function of number of used bases, $K$, is shown in Figure 2. 
%The number K is chosen as the minimum number such that the reconstructed background has a PSNR about 45 dB which results in a good reconstructed image quality.
We have found that K=10 leads to very good background representation for a variety of screen content images (with PSNR over 45dB), and additional bases do not lead to significant increase in the reconstruction quality.

Using this linear model, we need to solve the following optimization problem to derive model parameters:
\begin{gather*}
\{\alpha_1^*,...,\alpha_K^*\}= \argmin_{\alpha_1,...,\alpha_K} \| F(x,y)- \sum_{k=1}^K \alpha_k P_k(x,y)  \|_p
\end{gather*}
We can also look at the 1D version of this problem by converting the 2D blocks of size $N \times N$ into a vector of length $N^2$, denoted by $f$, by concatenating the columns and denoting $\sum_{k=1}^K \alpha_k P_k(x,y)$ as $ P\alpha$ where $P$ is a matrix of size $N^2\times K$ in which the k-th column corresponds to the vectorized version of $P_k(x,y)$ and $\alpha=[\alpha_1,...,\alpha_K]^\text{T}$. Then the problem can be formulated as: $\alpha^*= \argmin_{\alpha} \| f-P\alpha \|_p$

For the second question, we can use different distances between actual pixel values and approximated ones with the smooth model. As an example, by minimizing the l2 norm we will have the least-sqaure fitting problem.
%\begin{gather*}
%\alpha^*= \argmin_{\alpha} \| F-P\alpha \|_2 \Rightarrow \alpha= (P^T P)^{-1}P^T F
%\end{gather*}
The least square fitting suffers from the fact that the model parameters, $\alpha$, can be adversely affected by foreground pixels. In least-square fitting, by squaring the residuals, the larger residues will get larger weights in determining the model parameters. Because of that we propose to use least absolute deviation, which is more robust to outliers compared to least-square fitting and the model is less affected by outliers. Therefore we need to solve the following optimization problem:
\begin{gather}
\alpha^*= \argmin_{\alpha} \| f-P\alpha \|_1
\end{gather}
Least absolute deviation problem does not have a closed form solution but it can be solved with iterative algorithms. Different algorithms can be used to solve this problem, such as alternating direction method of multipliers (ADMM) \cite{ADMM}, iterative reweighted least square fitting \cite{IRLS} and linear programming. Here we use the ADMM algorithm. 
%This algorithm is explained in Section 2.1.

One alternative way to solve the second question is to use a robust regression approach to fit the smooth model into image blocks such that the model parameters are determined only using background pixels. One such a work is presented in \cite{ransac} where the author proposed to use RANSAC algorithm to fit the smooth model into the background pixels.

\subsection{ADMM formulation to solve least absolute deviation}
%ADMM is a variant of the augmented Lagrangian method which uses the partial update for dual variable. 
%It has been widely used in recent years, since it works for more general classes of problems than some other methods such as gradient descent (for example it works for cases where the objective function is not differentiable). 
To solve (1) with ADMM, we introduce the auxiliary variable $z= P\alpha-f$  and convert the original problem into the following form:
\begin{equation*}
\begin{aligned}
& \underset{z, \alpha}{\text{minimize}}
& & \| z \|_1 \\
& \text{subject to}
& & P\alpha-z=f.
\end{aligned}
\end{equation*}
Then we can use the following updates for each iteration in ADMM \cite{ADMM}:
%which is to alternatively update primal and dual variables as dual ascent, to solve this problem \cite{ADMM}:
\begin{flalign*}
& \alpha^{k+1}= (P^TP)^{-1}P^T(f+z^k-u^k) \\ 
& z^{k+1}= S_{1/{\rho}}(P\alpha^{k+1}-f+u^k) \\
& u^{k+1}= u^k+P\alpha^{k+1}-z^{k+1}-f
\end{flalign*}
where $u$ denotes the dual variable, $\rho$ is the augmented Lagrangian parameter and $S_{1/{\rho}}$ denotes soft-thresholding operator applied elementwise and is defined as:
\begin{gather*}
S_{1/{\rho}}(x)= \text{sign}(x) \text{max}(|x|-1/\rho,0)
\end{gather*}
%After some convergence criterion is satisfied, we will stop this algorithm.

\subsection{Segmentation Algorithm}
%After deriving the optimal parameters, $\alpha^*=[\alpha_1^*,...,\alpha_K^*]$, from least absolute deviation problem, we can reconstruct the intensity of each pixel at location $(x,y)$ using the smooth model $\hat{F}(x,y)= \sum_{k=1}^K \alpha^*_k P_k(x,y)$. Then each pixel will have a distortion $d(x,y)=|F(x,y)-\hat{F}(x,y)|$. Each pixel with a distortion less than a threshold $\epsilon_1$ will be considered as background and those with a distortion larger than the threshold will be considered as foreground. 
%This threshold could be either adaptive, depending on the image structure, or fixed.

We propose a segmentation algorithm which first checks if a block can be segmented using some simpler methods.
%and it goes to least absolute deviation only if the block cannot be segmented using those approaches.
These simple cases take care of two groups of blocks: completely flat block and smoothly varying background without foreground.
Completely flat blocks are those in which all pixels have the same value and are common in screen content images. Therefore they can be declared as background or foreground based on their neighboring blocks' background color. For these blocks, if we could find at least one neighbor block with a background color close enough to the current block's color (difference less than $\epsilon_2$), it would be segmented as background.
Smoothly varying background without foreground is a block in which the intensity of all pixels can be modeled well by the smooth function. Therefore we try to fit $K$ DCT basis to all pixels using least square fitting and if the intensity of all pixels can be predicted with distortion less than $\epsilon_3$, that block would be segmented entirely as background.
We will apply the least absolute fitting only if a block does not satisfy these two conditions. Furthermore, at the end of the least absolute fitting, we check the percentage of identified background pixels. If the percent is less than a threshold, we divide the block into 4 smaller blocks and repeat the process.
The overall segmentation algorithm is summarized below:
%\begin{center} \vspace{-0.5cm}
%\begin{tabular}{|p{0.98\linewidth}|}\hline % or any other width
%\rule{0pt}{5ex}% for more vertical space
%\vspace{-0.99cm}
\begin{enumerate}
\item If all pixels in the block have the same color intensity (i.e. it is completely flat block), declare the entire block as background or foreground as explained above. If not, go to the next step;
\item Perform least square fitting using the luminance values of all pixels. If all pixels can be predicted with an absolute error less than $\epsilon_3$, declare the entire block as background. If not, go to the next step;
\item Use least absolute deviation to fit a model to the luminance values of image block and find the absolute fitting error of all pixels using that model. Each pixel with a distortion less than a threshold $\epsilon_1$ will be considered as background, otherwise as foreground. If the percentage of background pixels is more than $\epsilon_4$ then stop, otherwise go to the next step;
\item Decompose current block of size $N \times N$ into 4 smaller blocks of size $\frac{N}{2} \times \frac{N}{2}$ and run the segmentation algorithm for all of them. Repeat until $N=8$.
\end{enumerate}

The above algorithm makes an initial decision based on the luminance component of a block only. At the end of this process, we further use least squares fitting to find a smooth model for  each of the two chrominance components (Cb and Cr) using the chrominance values at identified background pixels. If the fitting error for any color component is larger than $\epsilon_1$ for any pixel, that pixel is reclassified as foreground.

\section{Results}
\label{SectionIV}
%We have tested our algorithm on several test images. Most of them are selected from HEVC standard test sequences for screen content coding \cite{SCC}.
%They usually consist of texts, line and graphics overlaid on a smooth background. 
%We also show the result of the algorithm on a palmprint image to show its application for principal line extraction from palmprints.
To enable rigorous evaluation of different algorithms, we have generated an annotated dataset consisting of 332 image blocks of size $64\times 64$, extracted from sample frames from 5 HEVC test sequences for screen content coding. The ground truth foregrounds for these images are extracted manually.

Before showing the results let us discuss about parameters of our algorithm. 
In our implementation, the block size is chosen to be $N$=64 which is the same as the largest CU size in HEVC standard. The number of DCT basis functions, $K$, is chosen to be 10 based on the training images. %The maximum allowed distortion for a pixel to be considered background, $\epsilon_1$, is set to be 10 (out of the luminance range of 0 to 255). 
The other parameters are chosen as $\epsilon_1=10$, $\epsilon_2=10$, $\epsilon_3=3$ and $\epsilon_4=0.5$, which have been found to perform well on a training set. For ADMM algorithm, we have used the implementation by Stephen Boyd \cite{boyd}. The number of iteration is chosen to be 200 and the parameter $\rho$ is chosen as the default value, 1.

We compare the proposed algorithm with two algorithms; hierarchical k-means clustering in DjVu and shape primitive extraction and coding (SPEC) method. For SPEC, we have adapted the color number threshold and the shape primitive size threshold from the default value given in \cite{SPEC}  when necessary to give more satisfactory result. Furthermore, for blocks classified as text/graphics based on the color number, we segment the most frequent color and any similar color to it (i.e. colors which their distance from most frequent color is less than 10) in the current block as background and the rest as foreground.
%For active contour algorithm the MATLAB implementation \cite{active_imp} is used with the number of iteration set to 5000 so that the algorithm converges. The mask for active contour algorithm is set such that it contains all foreground objects.

\begin{figure*}[ht]
        \centering
        \vspace{-0.5cm}
        \begin{subfigure}[b]{0.18\textwidth}
       % \hspace{-1cm}
                \includegraphics[width=\textwidth]{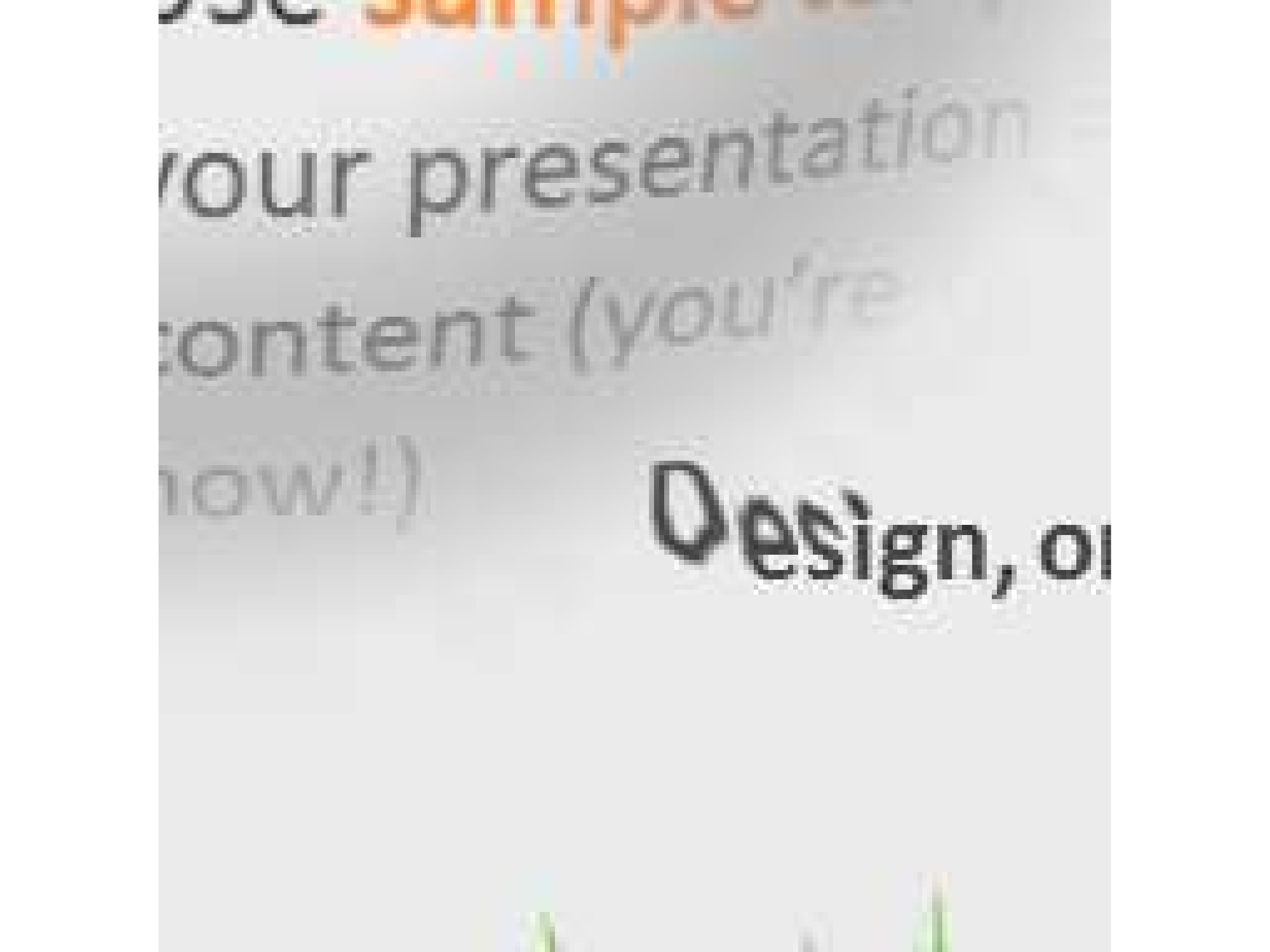}
                                \vspace{-0.5cm}
          \hspace{-2.5cm}    
        \end{subfigure}%
        ~ %add desired spacing between images, e. g. ~, \quad, \qquad, \hfill etc.
        \begin{subfigure}[b]{0.18\textwidth}
       % \hspace{-2cm}
                \includegraphics[width=\textwidth]{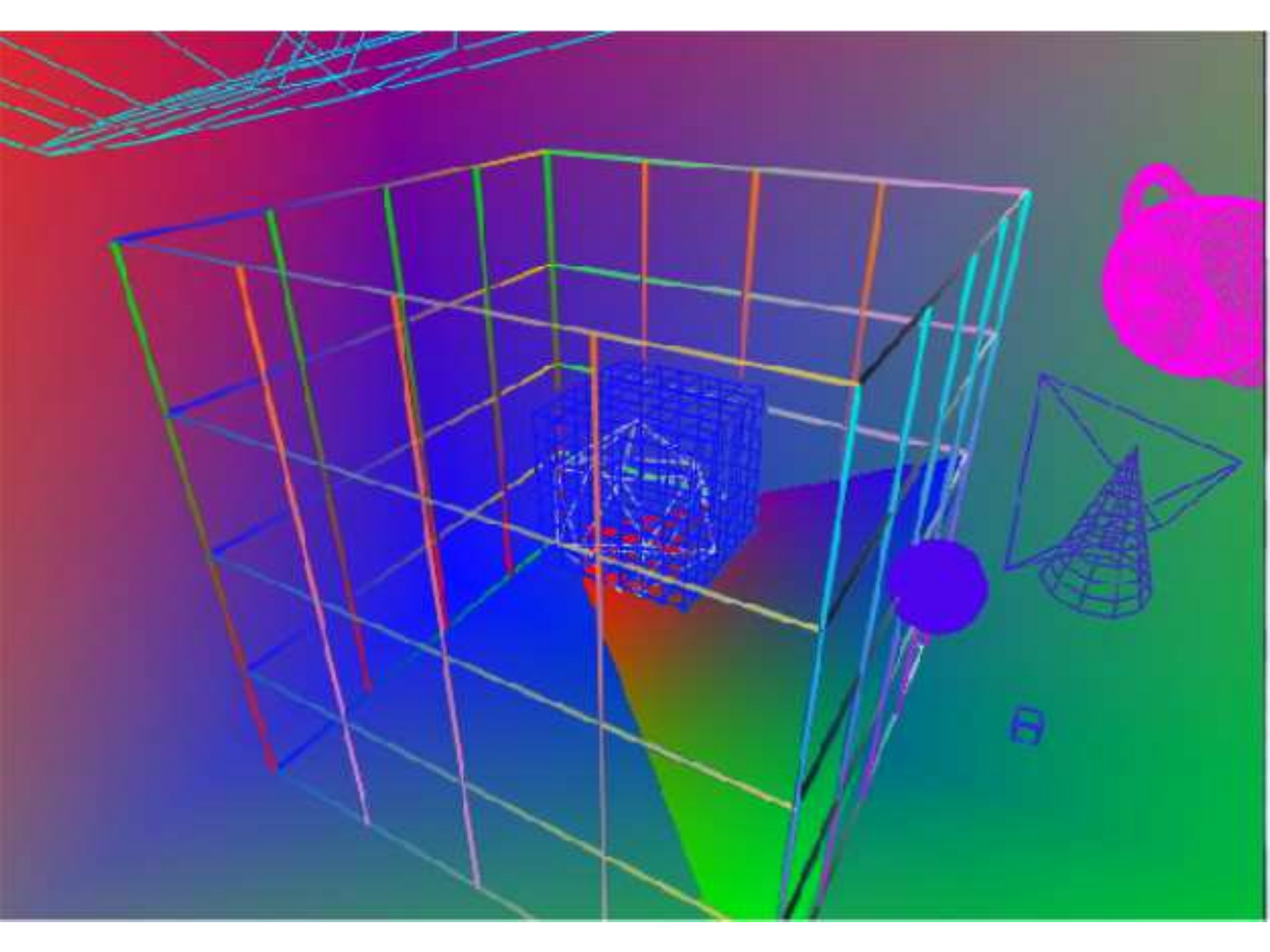}
                \vspace{-0.5cm}
            \hspace{-3cm} 
        \end{subfigure}%
        ~ %add desired spacing between images, e. g. ~, \quad, \qquad, \hfill etc.
        \begin{subfigure}[b]{0.18\textwidth}
       % \hspace{-2cm}
                \includegraphics[width=\textwidth]{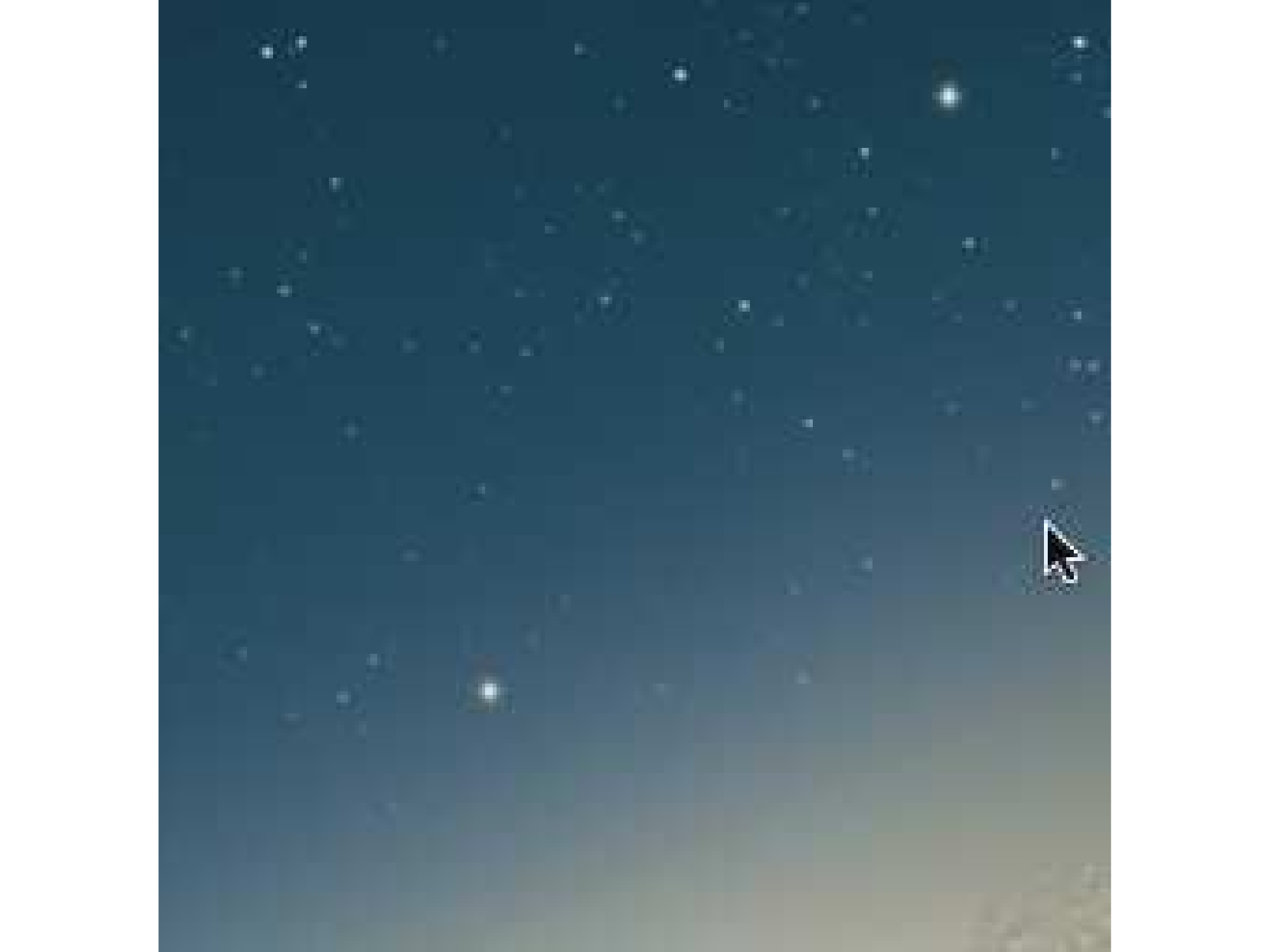}
                \vspace{-0.45cm}
            \hspace{-3cm} 
        \end{subfigure}%    
        \begin{subfigure}[b]{0.18\textwidth}
			~ %add desired spacing between images, e. g. ~, \quad, \qquad, \hfill etc.
                \includegraphics[width=\textwidth]{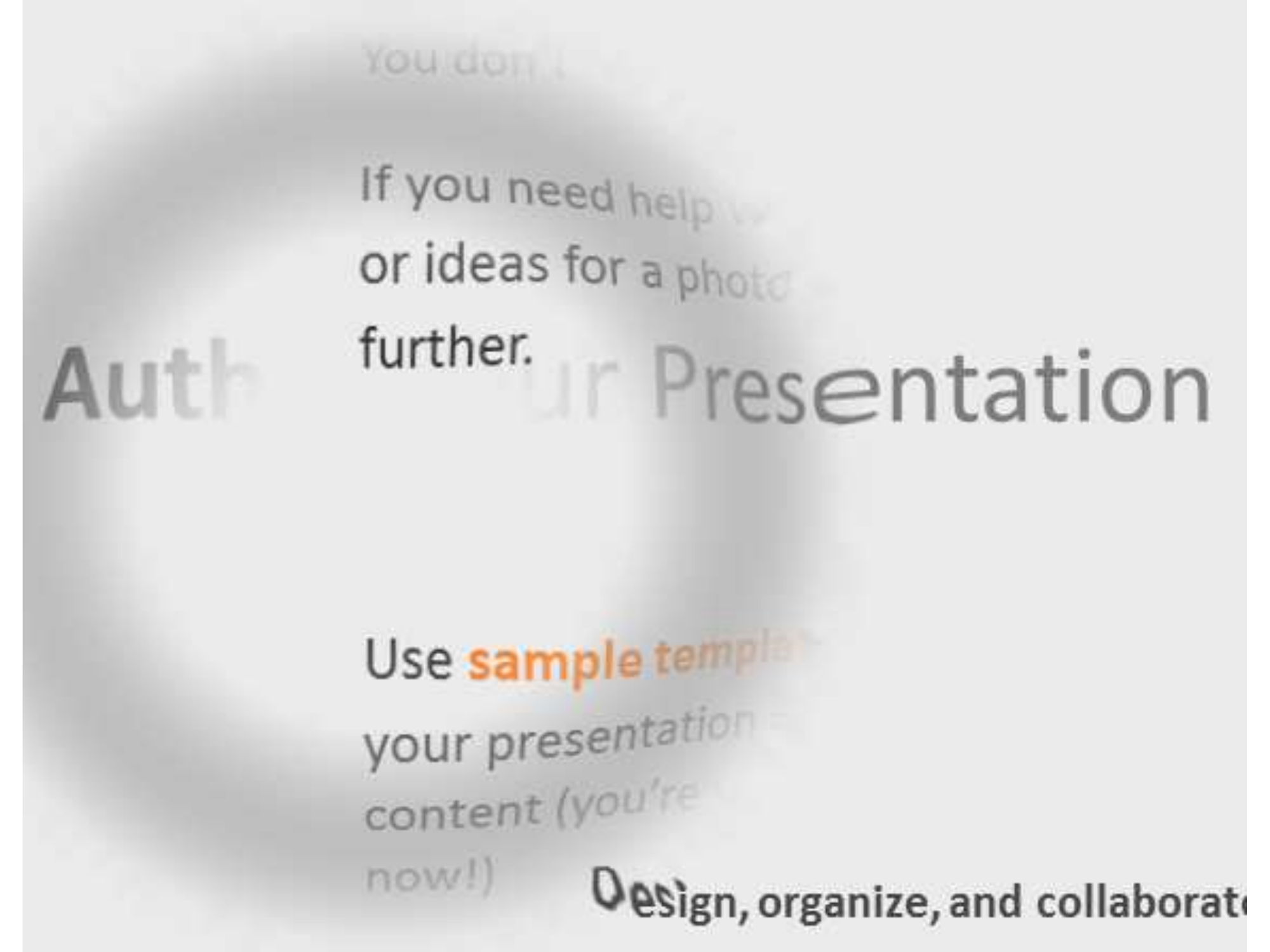}
                \vspace{-0.45cm}
            \hspace{-5cm} 
        \end{subfigure}%    
        \begin{subfigure}[b]{0.18\textwidth}
      %  \hspace{-3cm}
                \includegraphics[width=\textwidth]{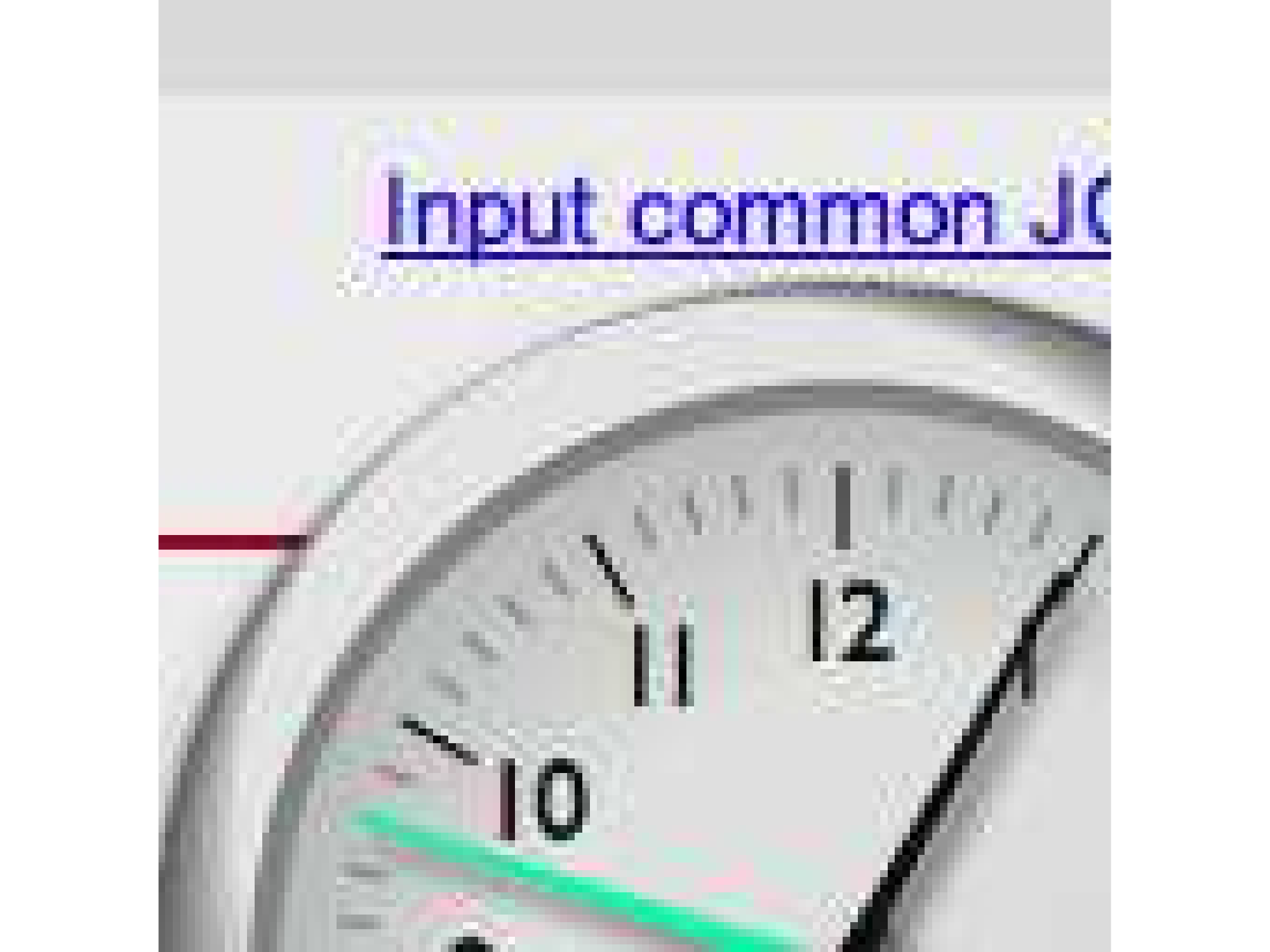}
                 \vspace{-0.45cm}
              \hspace{-4.8cm}
        \end{subfigure}
         \\[1ex]
        \begin{subfigure}[b]{0.18\textwidth}
       % \hspace{-1cm}
                \includegraphics[width=\textwidth]{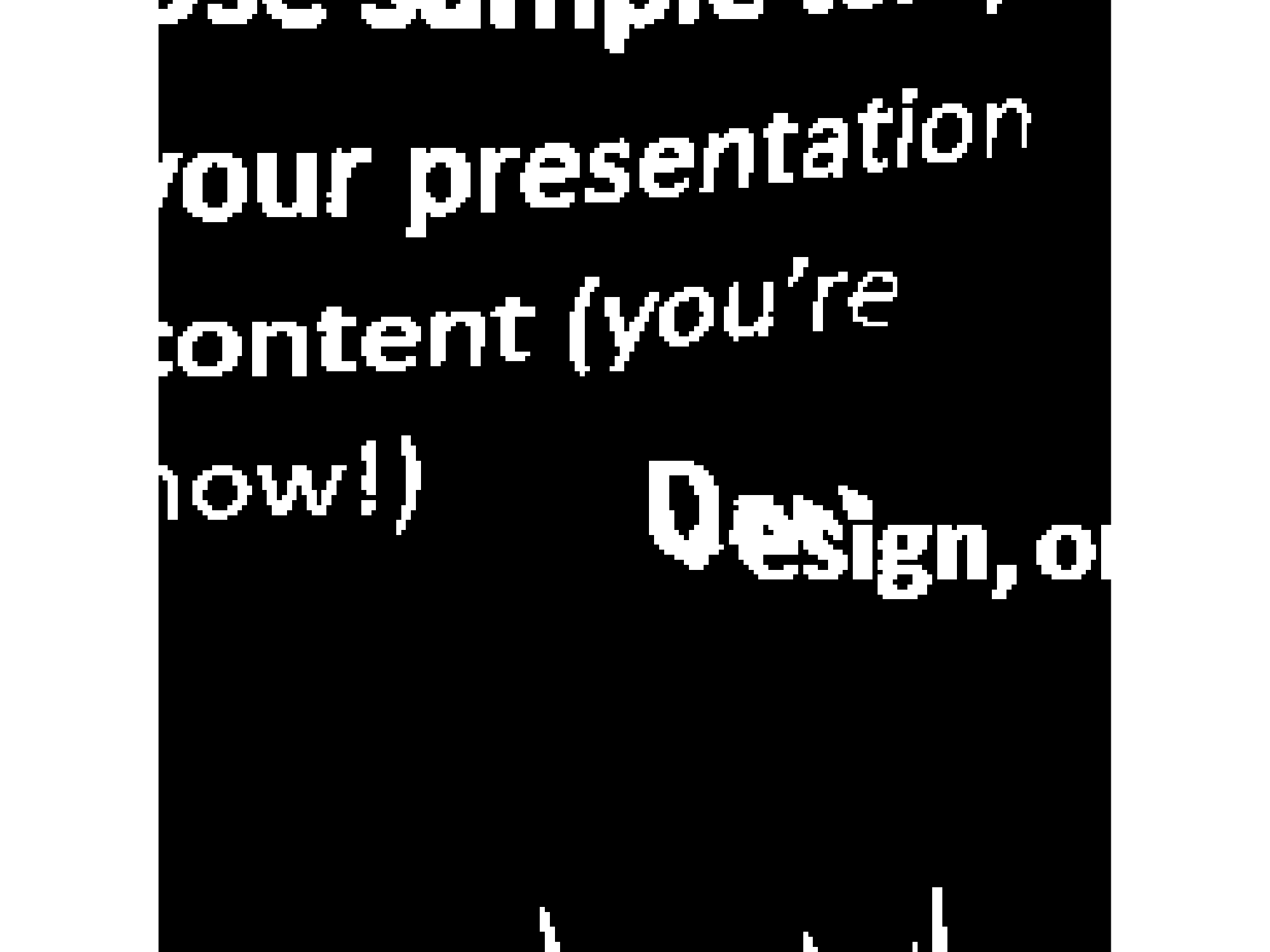}
                                \vspace{-0.5cm}
          \hspace{-2.5cm}    
        \end{subfigure}%
        ~ %add desired spacing between images, e. g. ~, \quad, \qquad, \hfill etc.
        \begin{subfigure}[b]{0.18\textwidth}
       % \hspace{-2cm}
                \includegraphics[width=\textwidth]{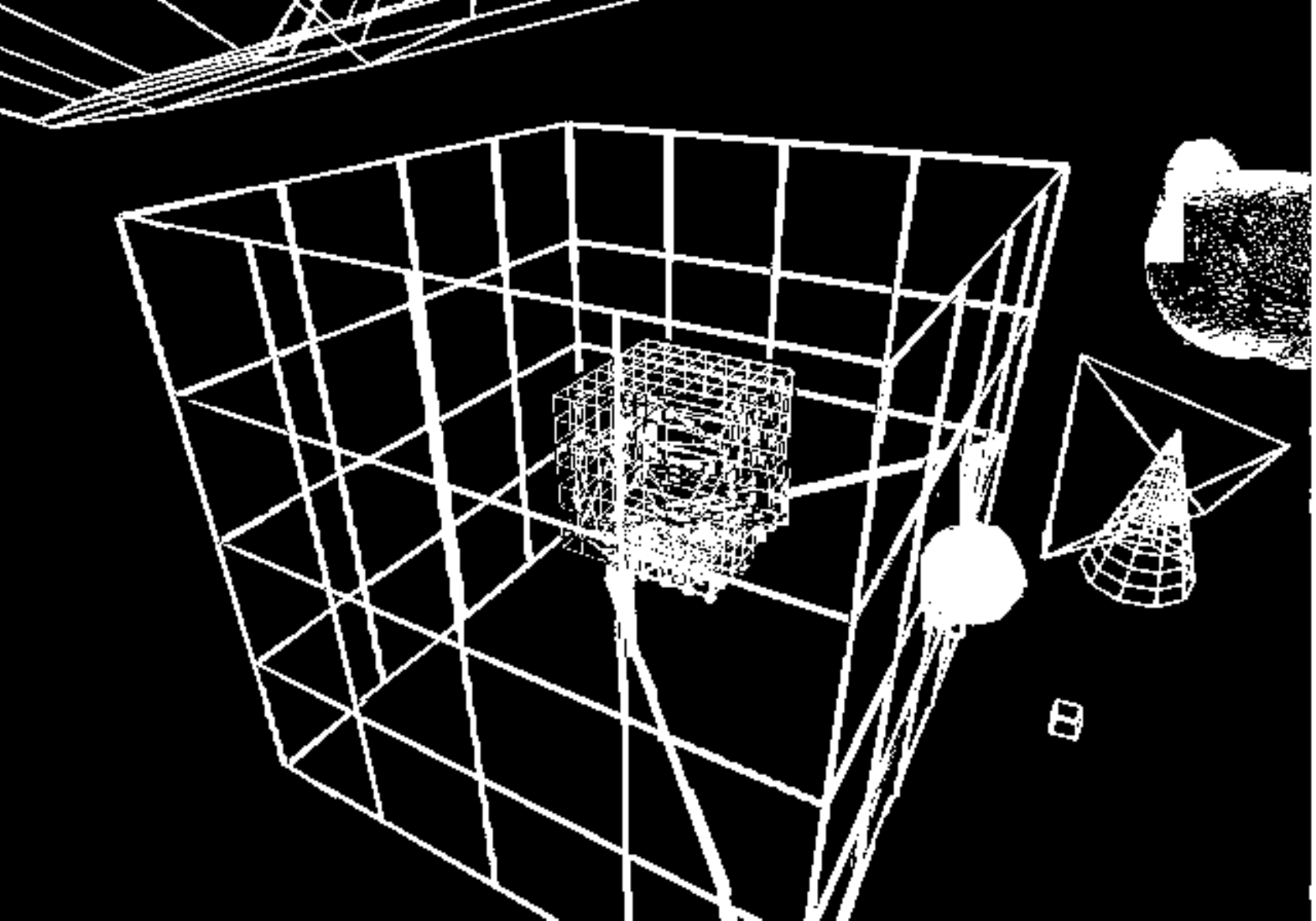}
                \vspace{-0.5cm}
            \hspace{-3cm} 
        \end{subfigure}%
        ~ %add desired spacing between images, e. g. ~, \quad, \qquad, \hfill etc.
        \begin{subfigure}[b]{0.18\textwidth}
       % \hspace{-2cm}
                \includegraphics[width=\textwidth]{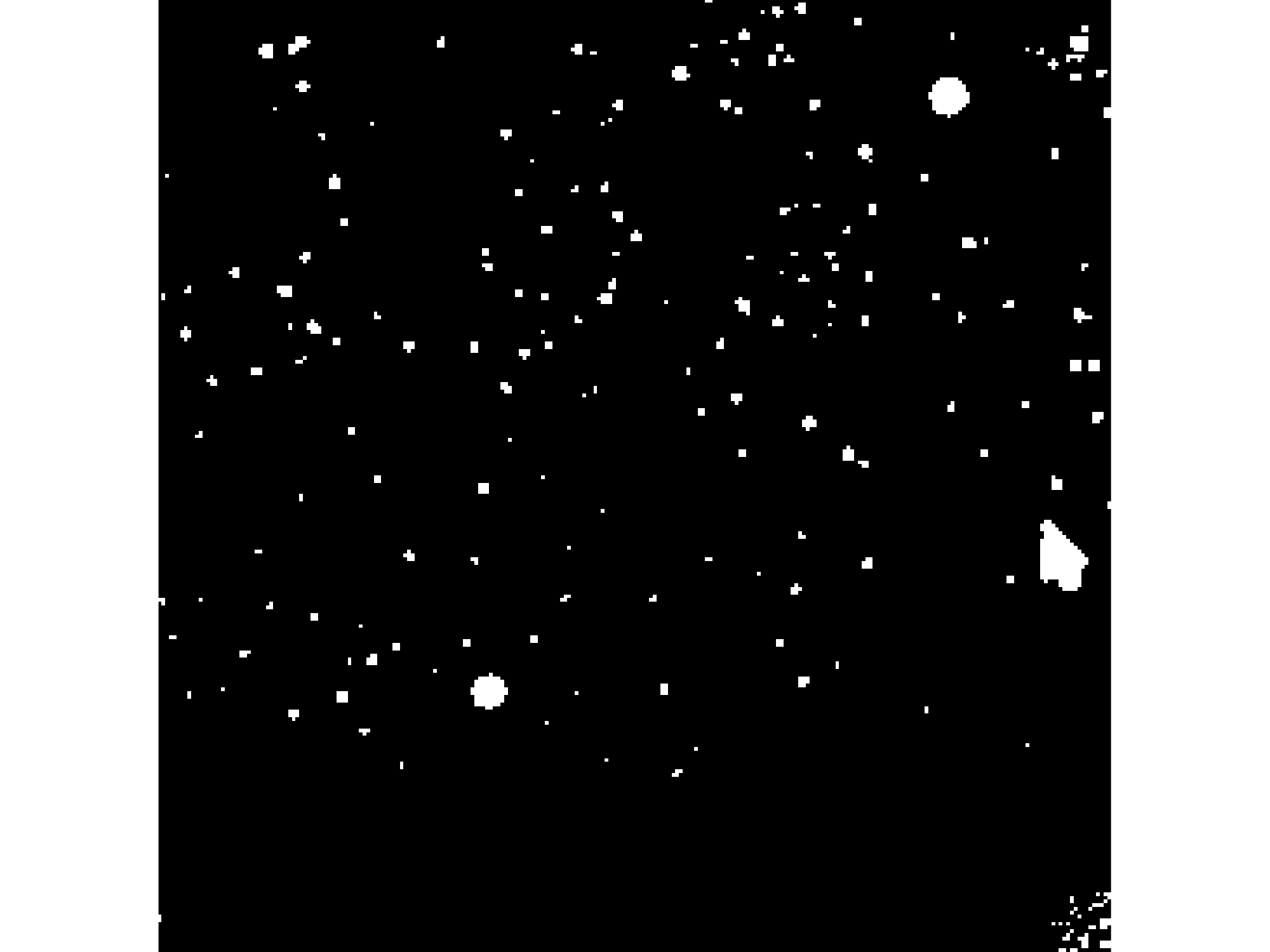}
                \vspace{-0.45cm}
            \hspace{-3cm} 
        \end{subfigure}%      
        \begin{subfigure}[b]{0.18\textwidth}
			~ %add desired spacing between images, e. g. ~, \quad, \qquad, \hfill etc.
                \includegraphics[width=\textwidth]{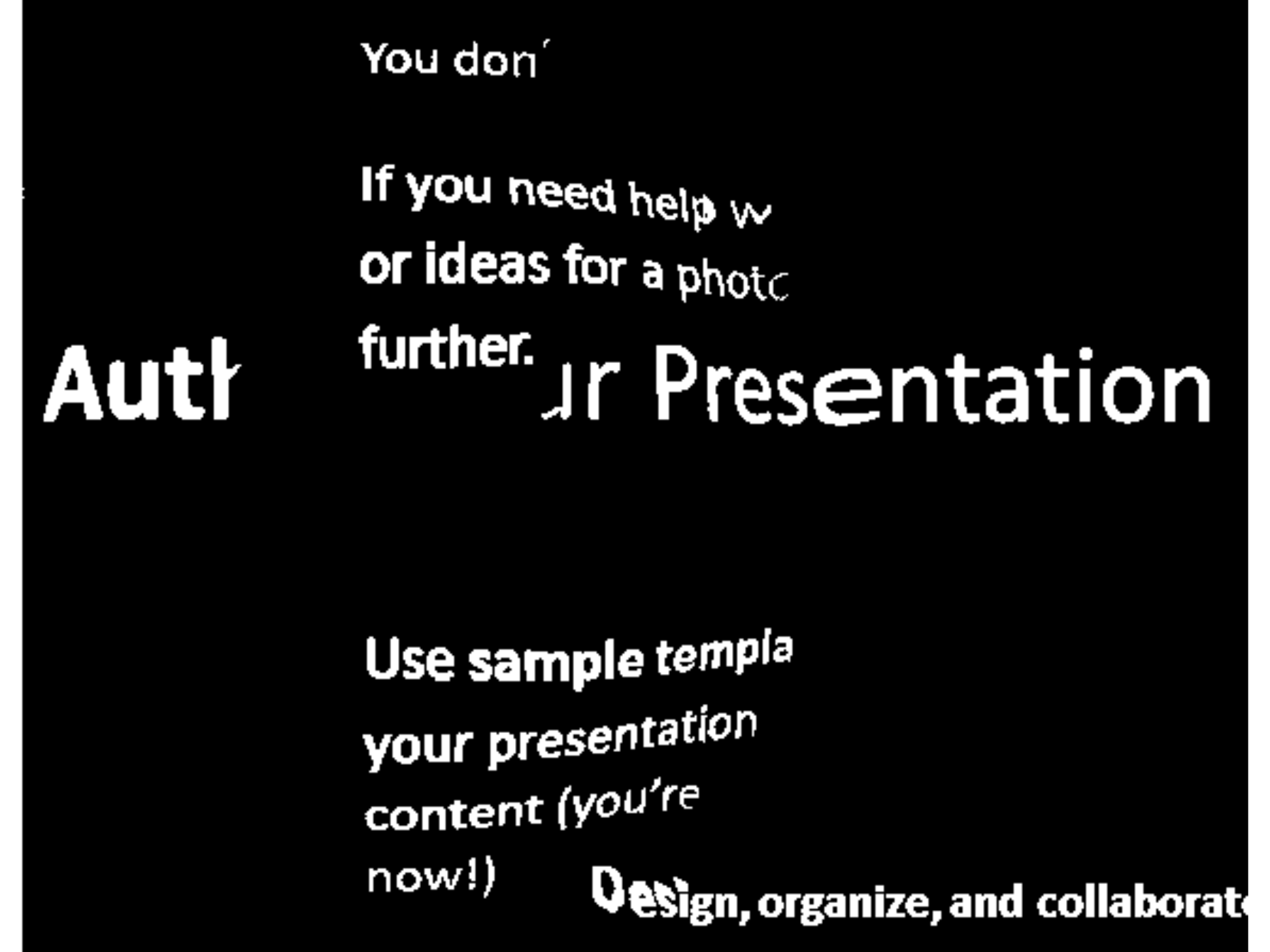}
                \vspace{-0.45cm}
            \hspace{-3cm} 
        \end{subfigure}%            
        \begin{subfigure}[b]{0.18\textwidth}
      %  \hspace{-3cm}
                \includegraphics[width=\textwidth]{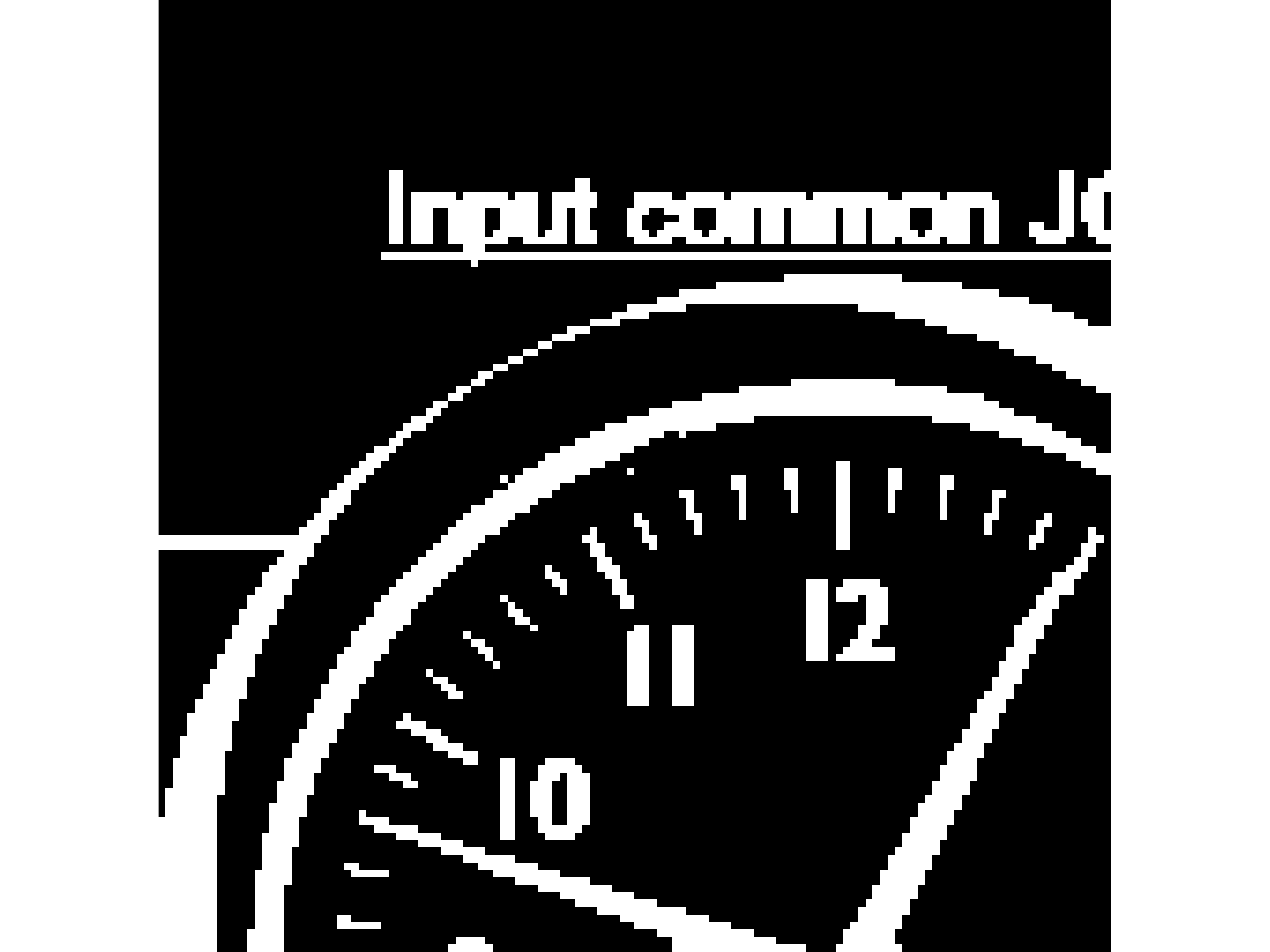}
                 \vspace{-0.45cm}
              \hspace{-4.8cm}
        \end{subfigure} \\[1ex]
        \begin{subfigure}[b]{0.18\textwidth}
       % \hspace{-1cm}
                \includegraphics[width=\textwidth]{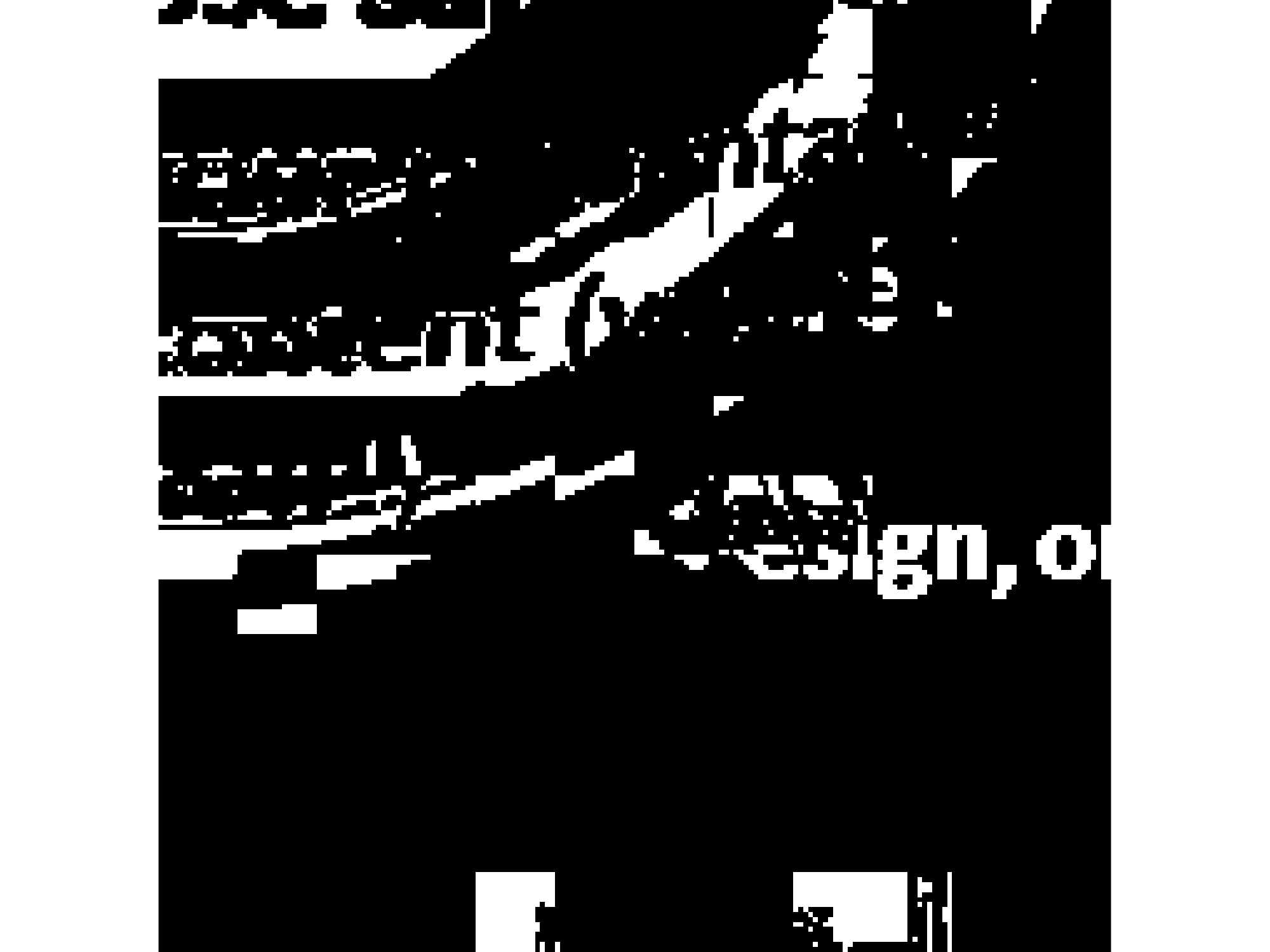}
                                \vspace{-0.5cm}
          \hspace{-2.5cm}    
        \end{subfigure}%
        ~ %add desired spacing between images, e. g. ~, \quad, \qquad, \hfill etc.
        \begin{subfigure}[b]{0.18\textwidth}
       % \hspace{-2cm}
                \includegraphics[width=\textwidth]{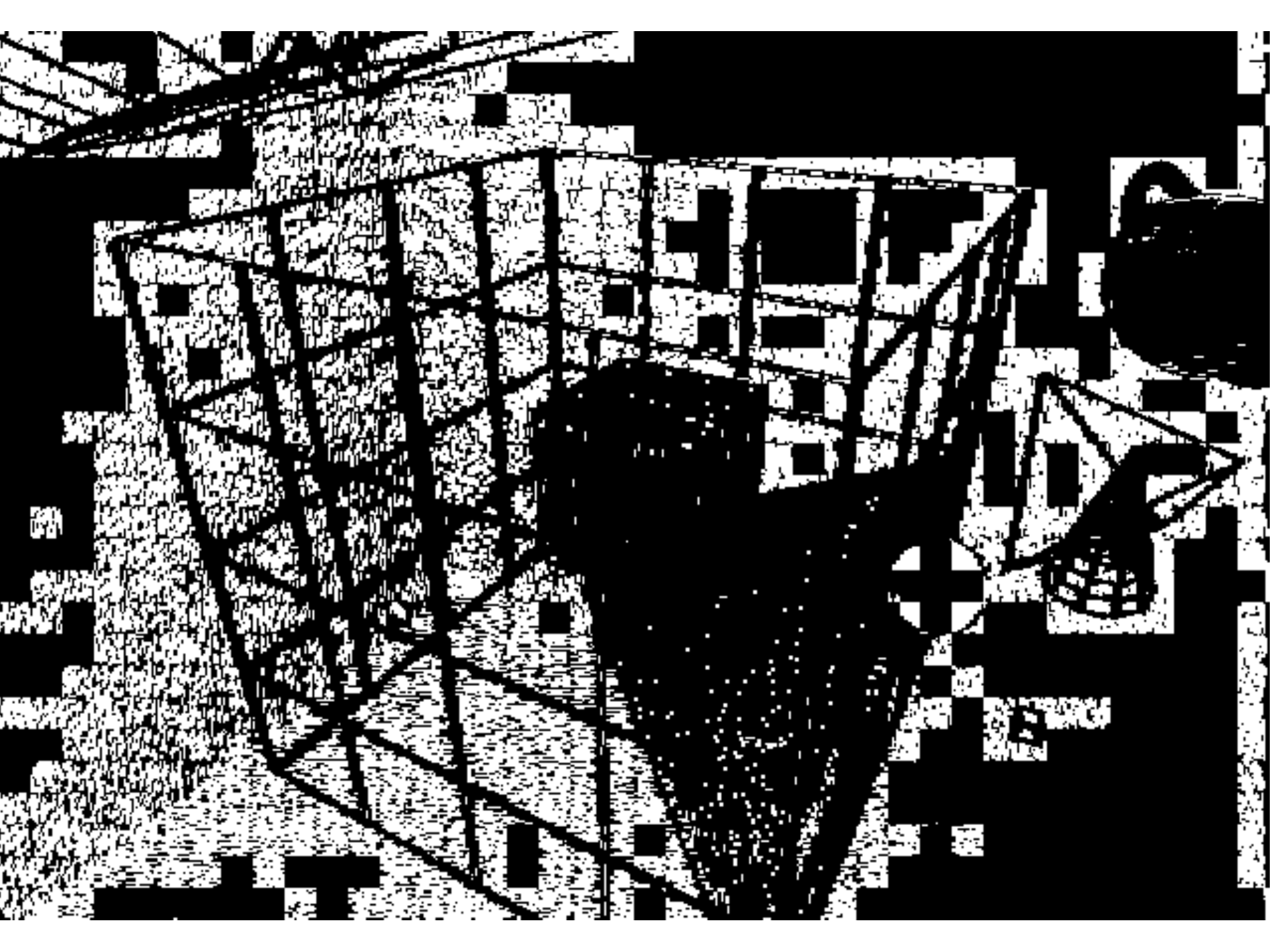}
                \vspace{-0.5cm}
            \hspace{-3cm} 
        \end{subfigure}%
        ~ %add desired spacing between images, e. g. ~, \quad, \qquad, \hfill etc.
        \begin{subfigure}[b]{0.18\textwidth}
       % \hspace{-2cm}
                \includegraphics[width=\textwidth]{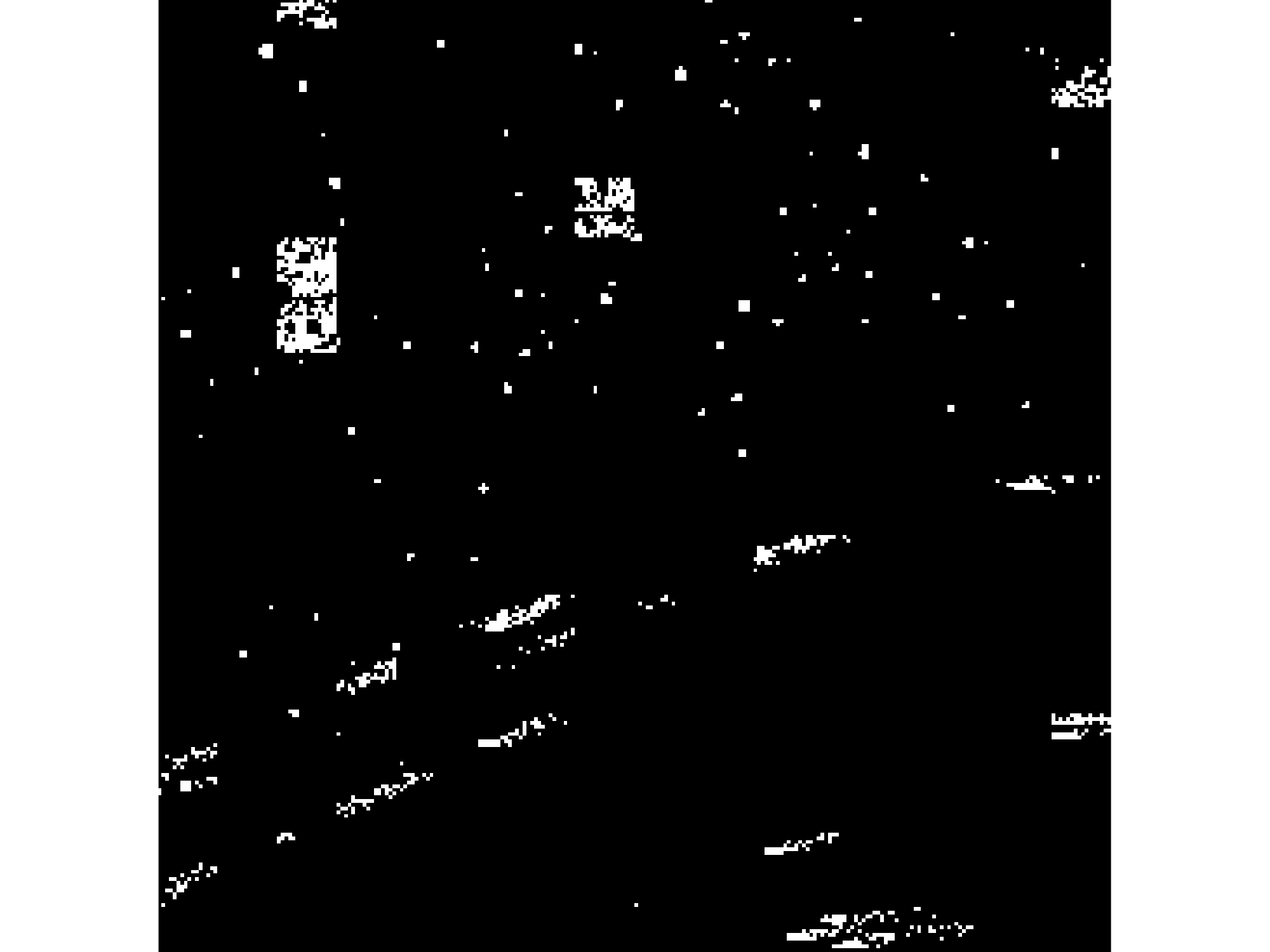}
                \vspace{-0.45cm}
            \hspace{-3cm} 
        \end{subfigure}%      
        \begin{subfigure}[b]{0.18\textwidth}
			~ %add desired spacing between images, e. g. ~, \quad, \qquad, \hfill etc.
                \includegraphics[width=\textwidth]{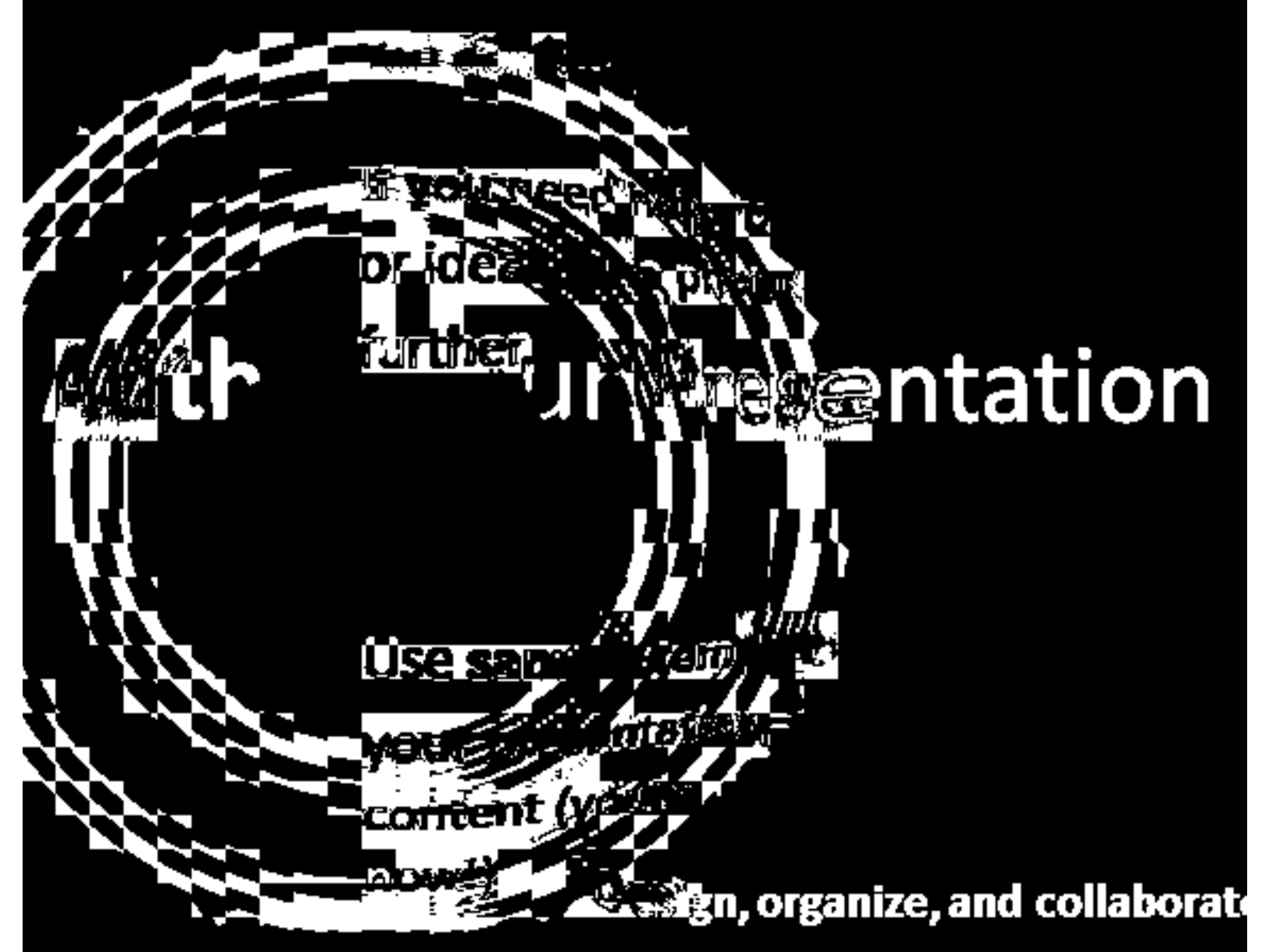}
                \vspace{-0.45cm}
            \hspace{-3cm} 
        \end{subfigure}%            
        \begin{subfigure}[b]{0.18\textwidth}
      %  \hspace{-3cm}
                \includegraphics[width=\textwidth]{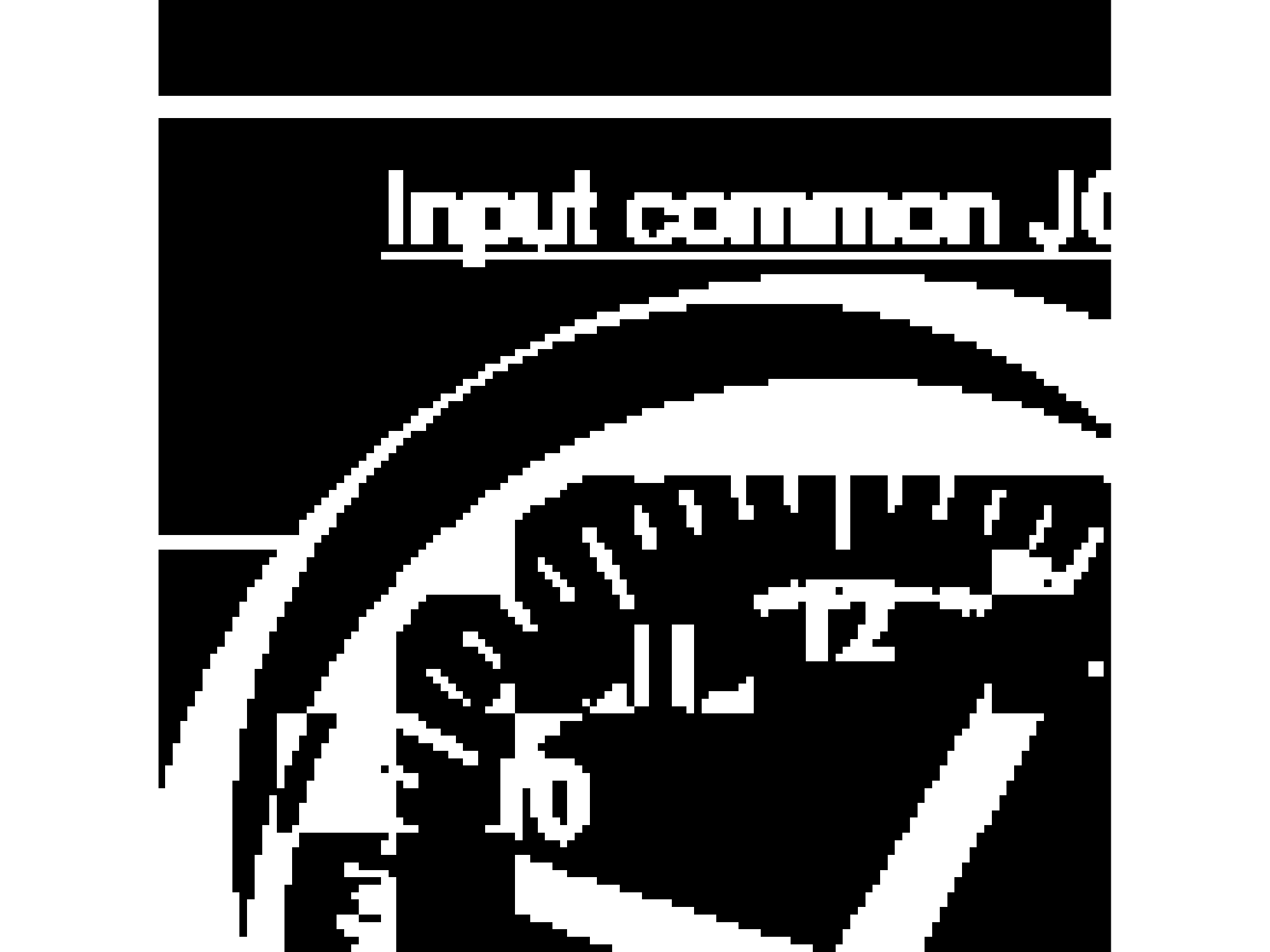}
                 \vspace{-0.45cm}
              \hspace{-4.8cm}
        \end{subfigure} \\[1ex]
        \begin{subfigure}[b]{0.18\textwidth}
       % \hspace{-1cm}
                \includegraphics[width=\textwidth]{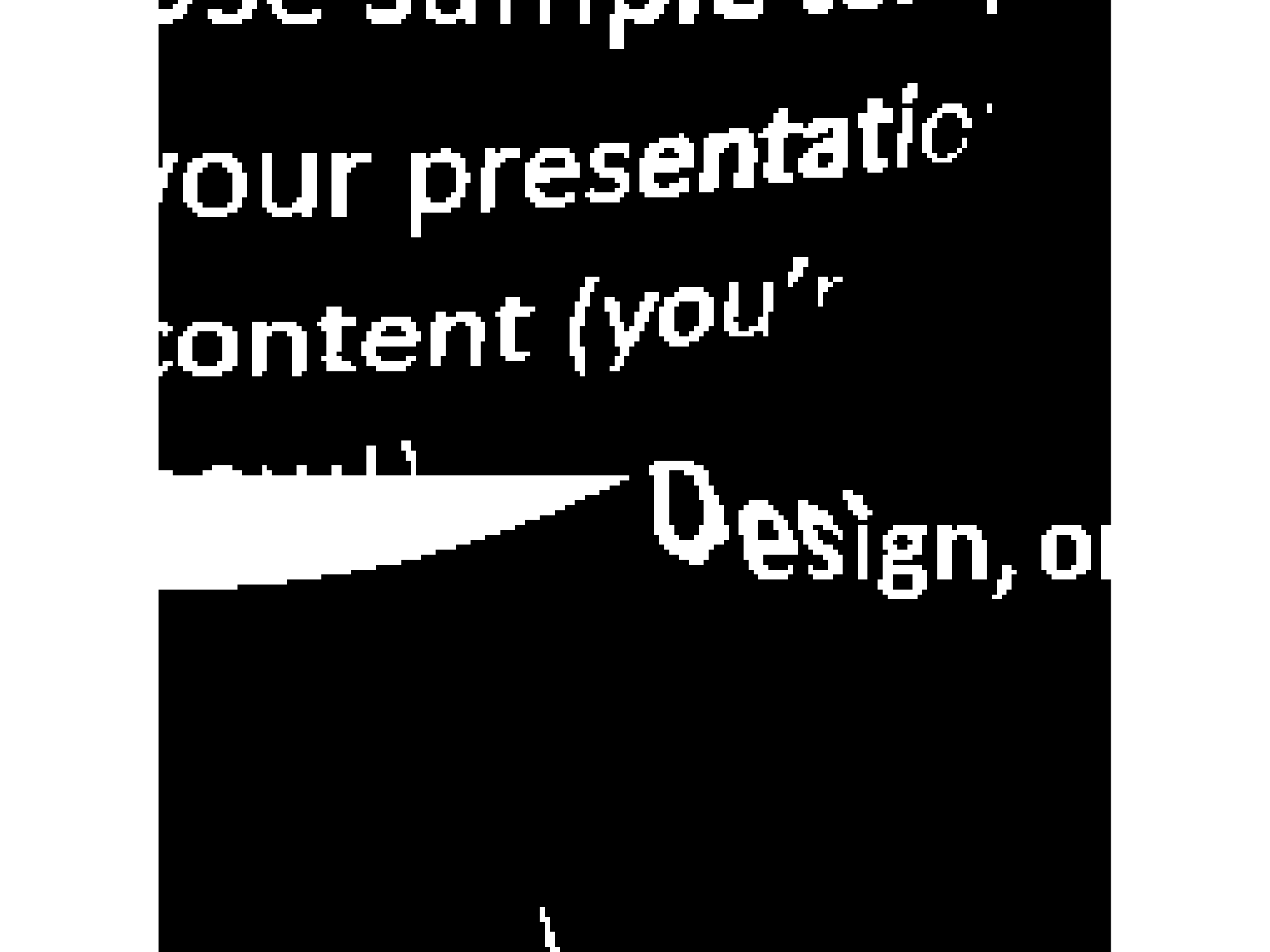}
                                \vspace{-0.5cm}
          \hspace{-2.5cm}    
        \end{subfigure}%
        ~ %add desired spacing between images, e. g. ~, \quad, \qquad, \hfill etc.
        \begin{subfigure}[b]{0.18\textwidth}
       % \hspace{-2cm}
                \includegraphics[width=\textwidth]{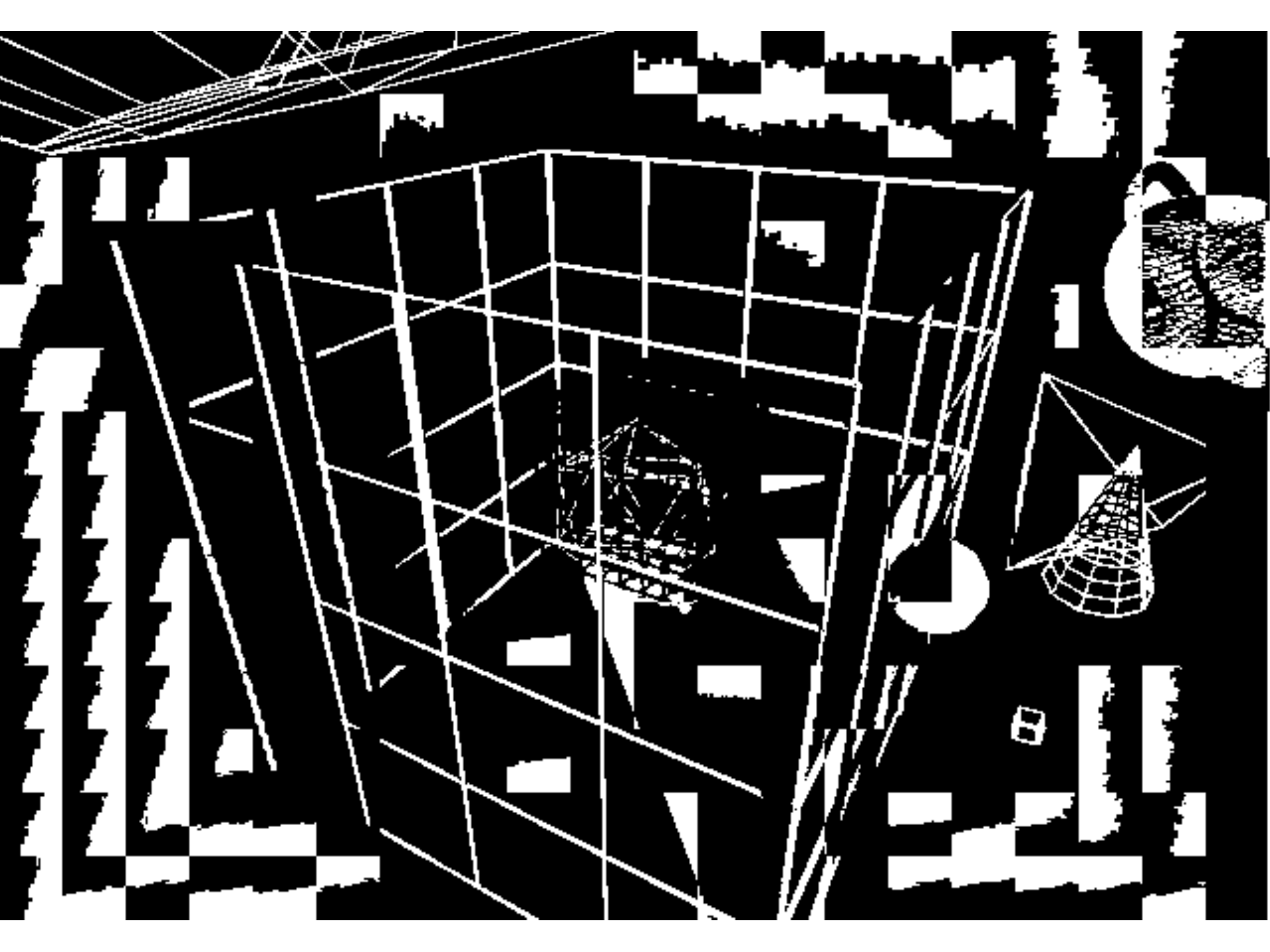}
                \vspace{-0.5cm}
            \hspace{-3cm} 
        \end{subfigure}%
        ~ %add desired spacing between images, e. g. ~, \quad, \qquad, \hfill etc.
        \begin{subfigure}[b]{0.18\textwidth}
       % \hspace{-2cm}
                \includegraphics[width=\textwidth]{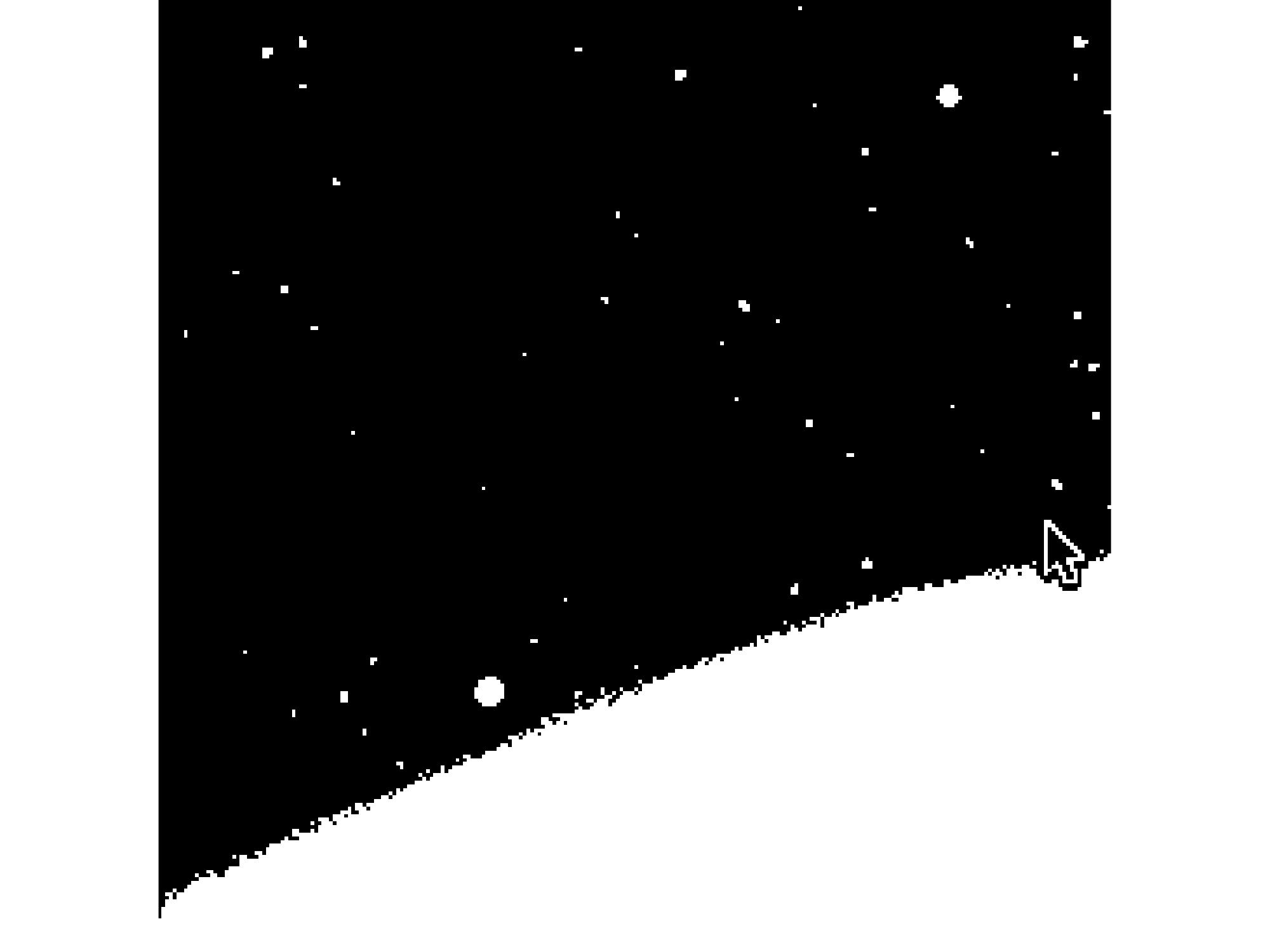}
                \vspace{-0.45cm}
            \hspace{-3cm} 
        \end{subfigure}%   
        \begin{subfigure}[b]{0.18\textwidth}
			~ %add desired spacing between images, e. g. ~, \quad, \qquad, \hfill etc.
                \includegraphics[width=\textwidth]{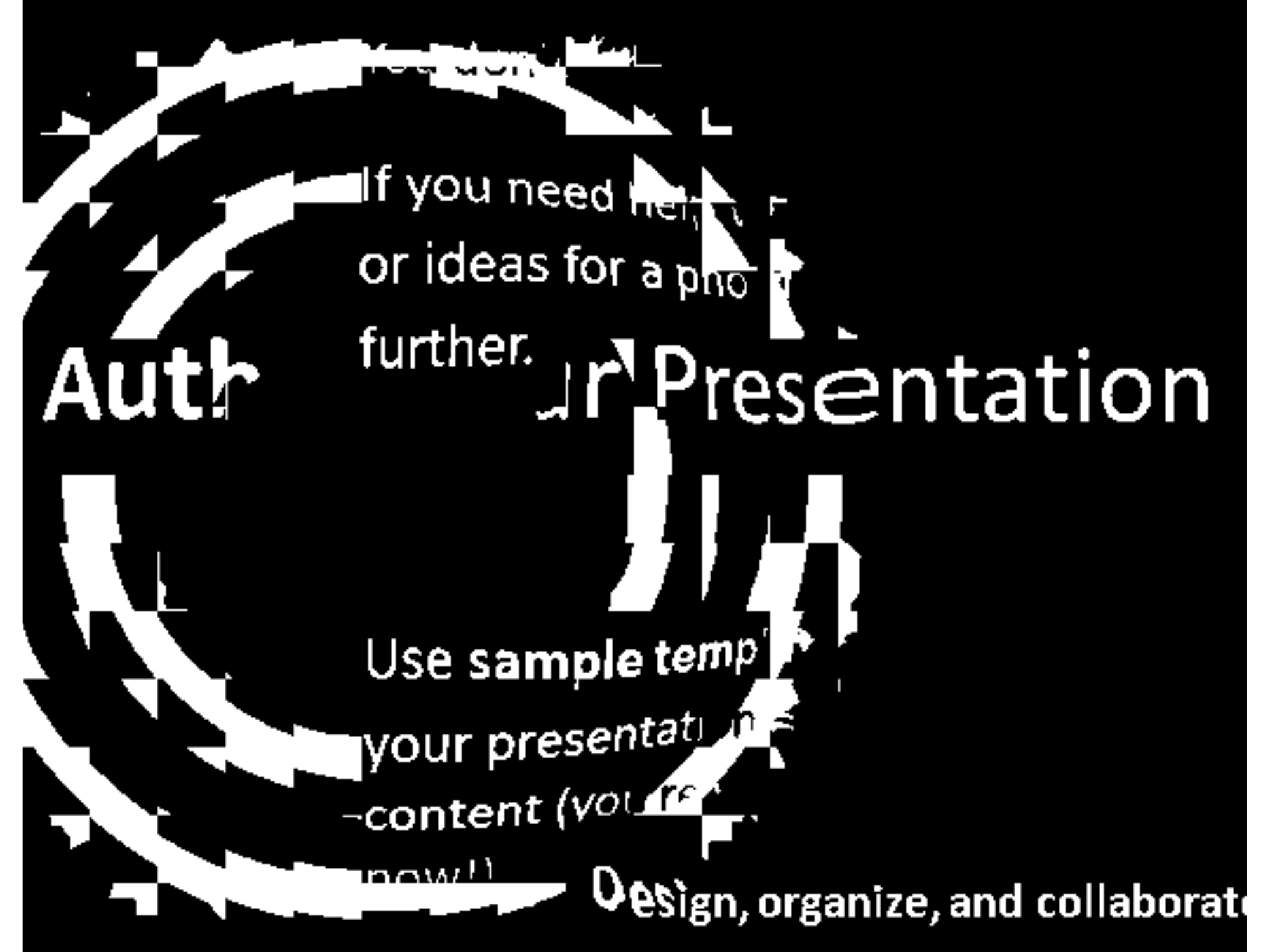}
                \vspace{-0.45cm}
            \hspace{-3cm} 
        \end{subfigure}%            
        \begin{subfigure}[b]{0.18\textwidth}
      %  \hspace{-3cm}
                \includegraphics[width=\textwidth]{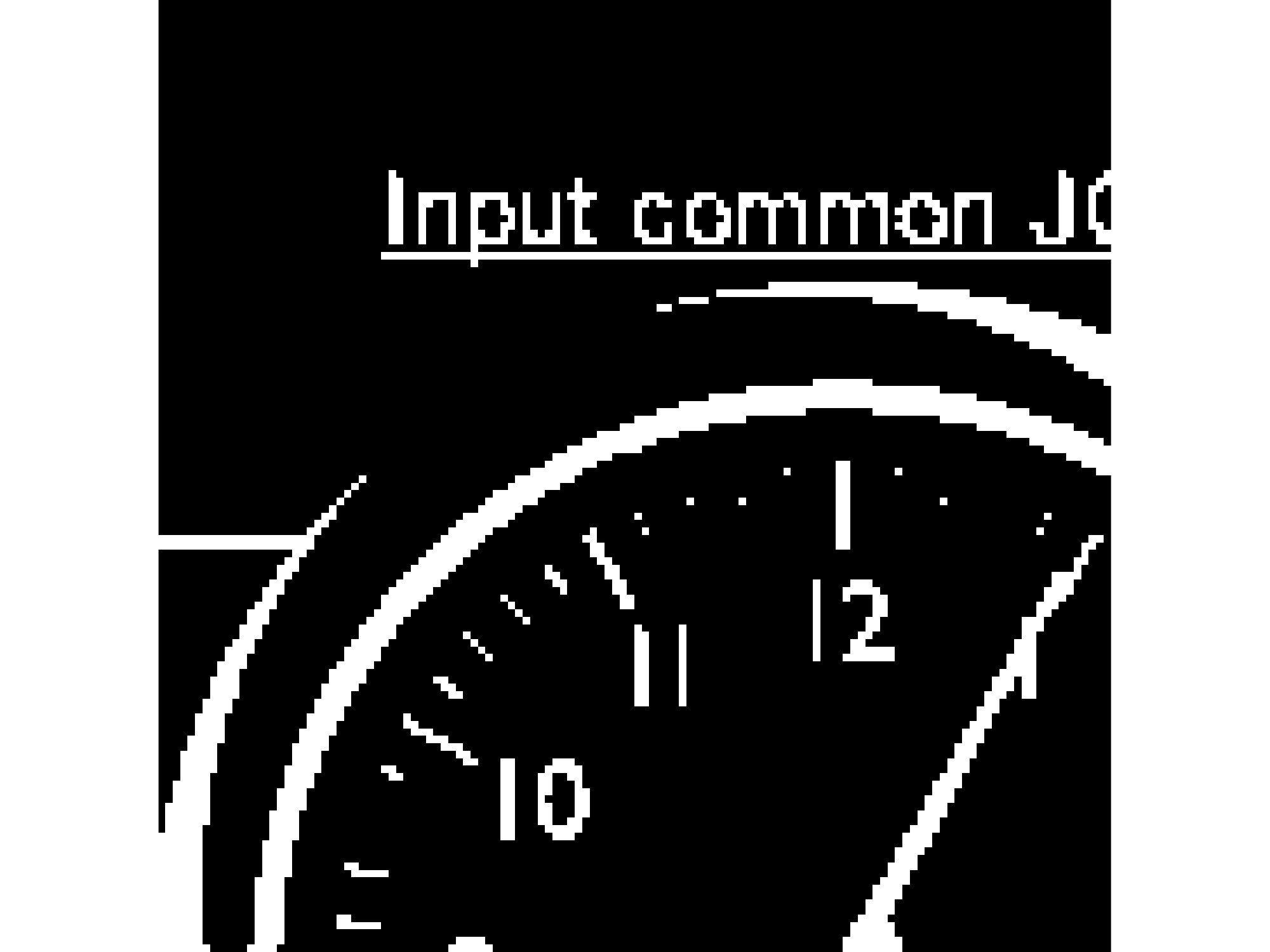}
                 \vspace{-0.45cm}
              \hspace{-4.8cm}
        \end{subfigure} \\[1ex]        
        \begin{subfigure}[b]{0.18\textwidth}
       % \hspace{-1cm}
                \includegraphics[width=\textwidth]{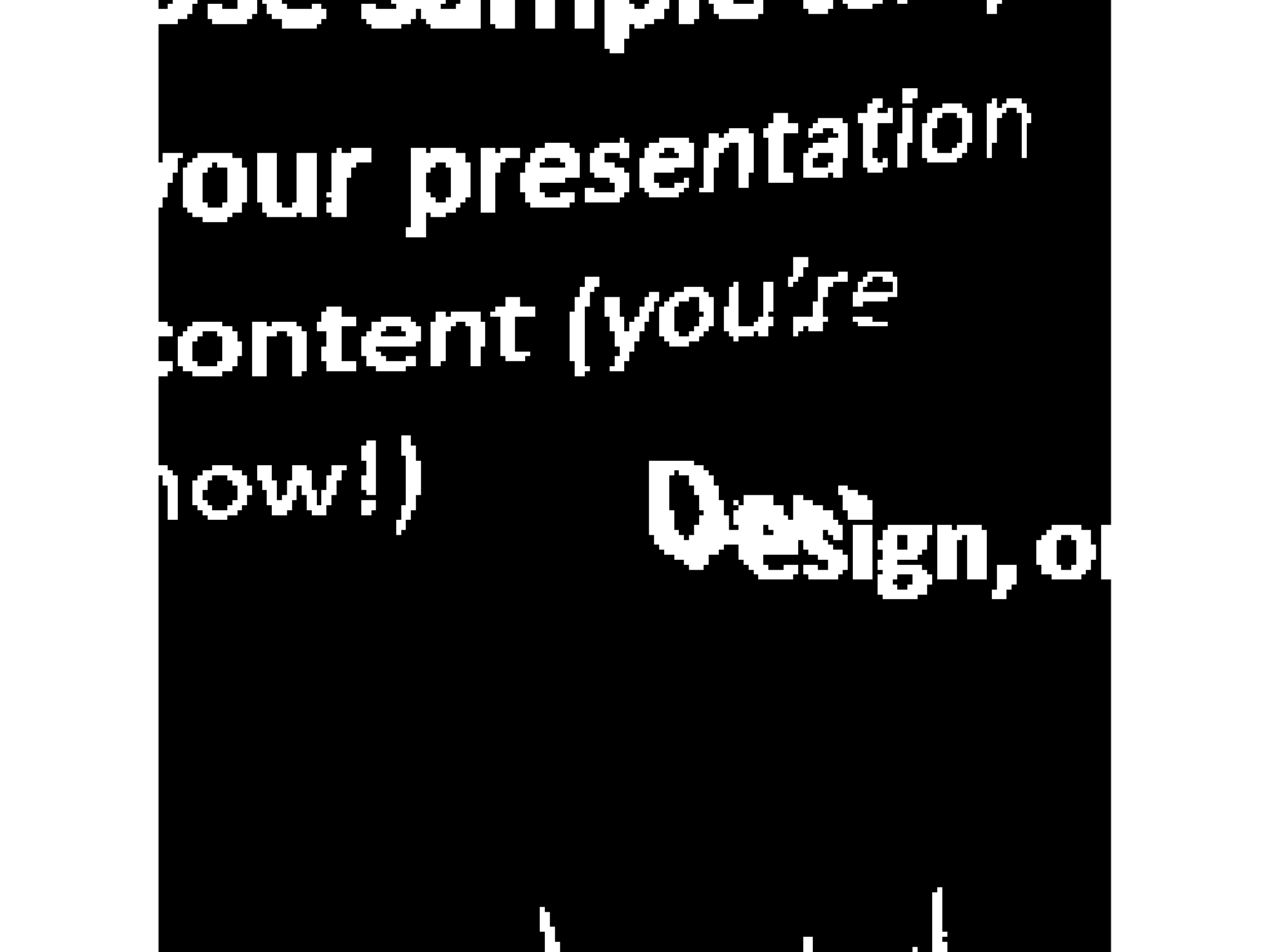}
                                \vspace{-0.5cm}
          \hspace{-2.5cm}    
        \end{subfigure}%
        ~ %add desired spacing between images, e. g. ~, \quad, \qquad, \hfill etc.
        \begin{subfigure}[b]{0.18\textwidth}
       % \hspace{-2cm}
                \includegraphics[width=\textwidth]{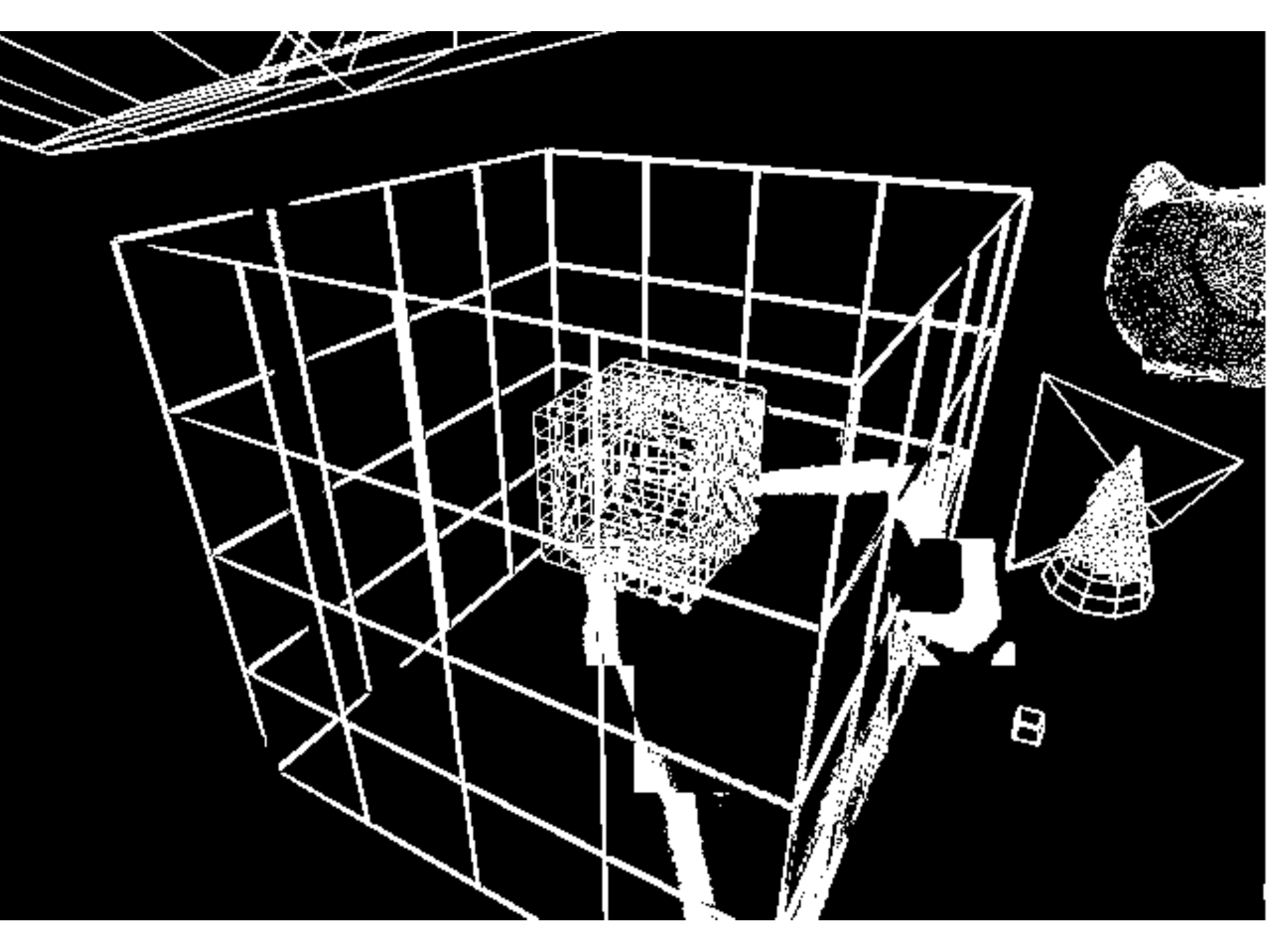}
                \vspace{-0.5cm}
            \hspace{-3cm} 
        \end{subfigure}%
        ~ %add desired spacing between images, e. g. ~, \quad, \qquad, \hfill etc.
        \begin{subfigure}[b]{0.18\textwidth}
       % \hspace{-2cm}
                \includegraphics[width=\textwidth]{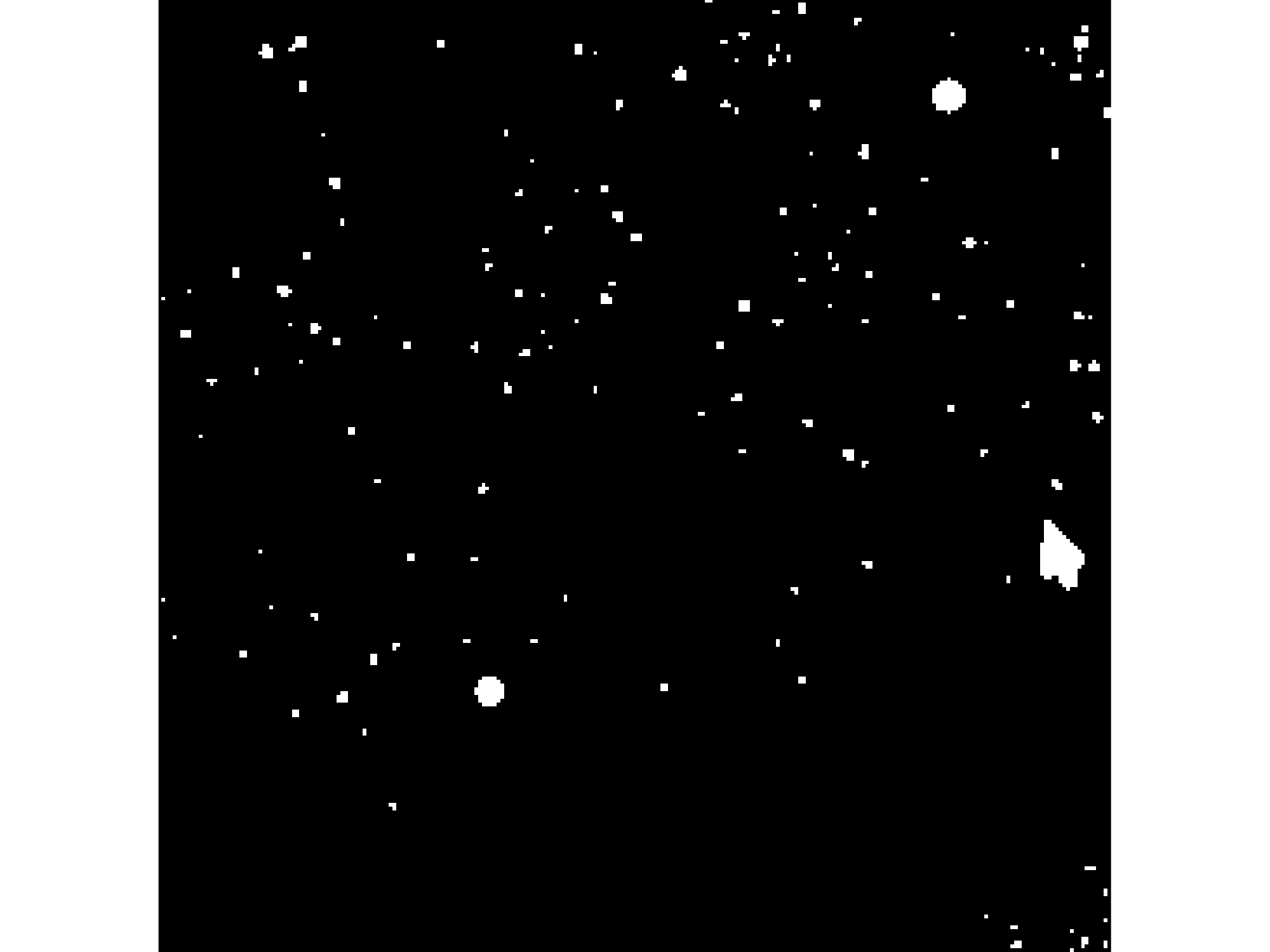}
                \vspace{-0.45cm}
            \hspace{-3cm} 
        \end{subfigure}%
        \begin{subfigure}[b]{0.18\textwidth}
			~ %add desired spacing between images, e. g. ~, \quad, \qquad, \hfill etc.
                \includegraphics[width=\textwidth]{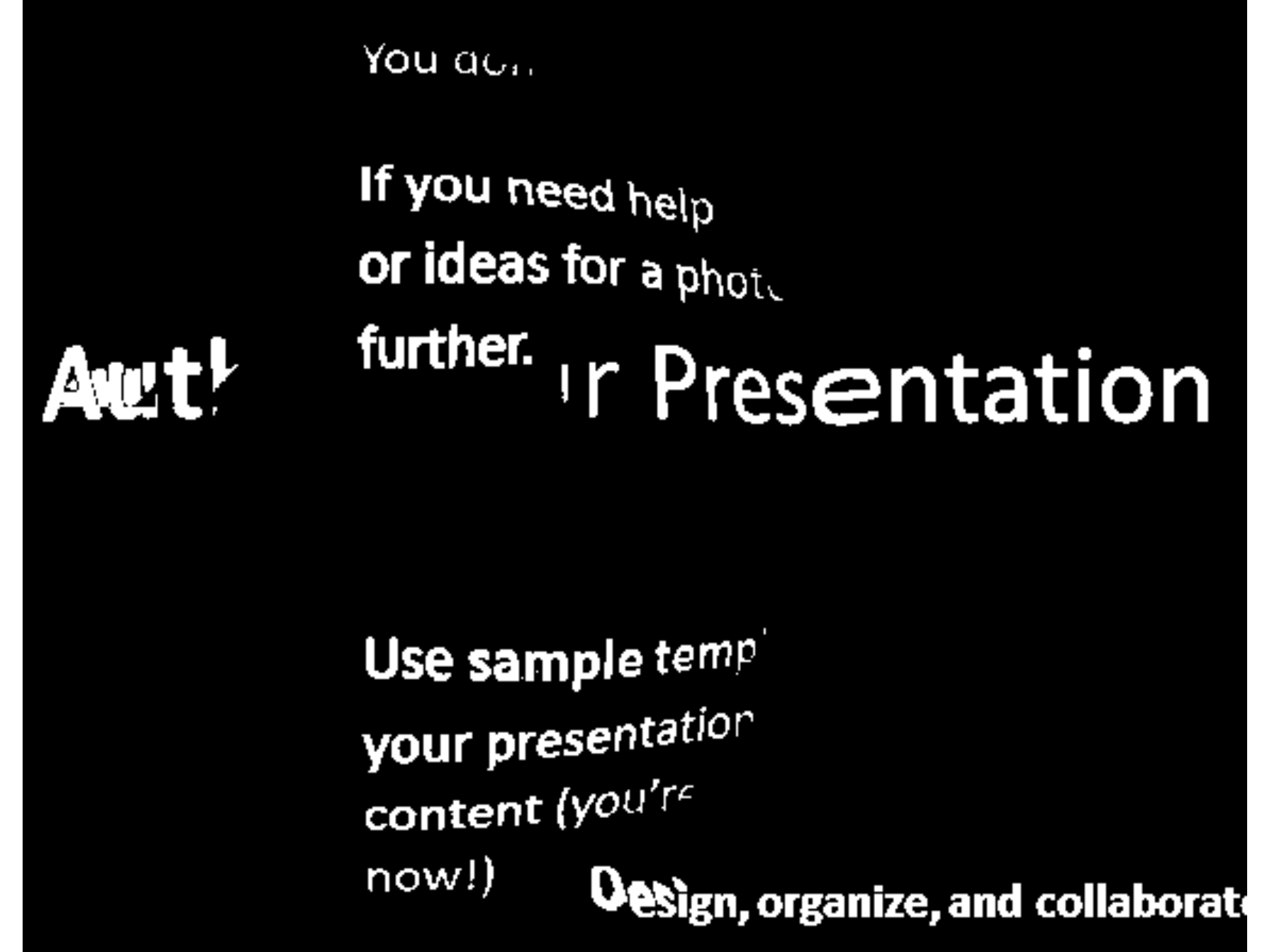}
                \vspace{-0.45cm}
            \hspace{-3cm} 
        \end{subfigure}%                  
        \begin{subfigure}[b]{0.18\textwidth}
      %  \hspace{-3cm}
                \includegraphics[width=\textwidth]{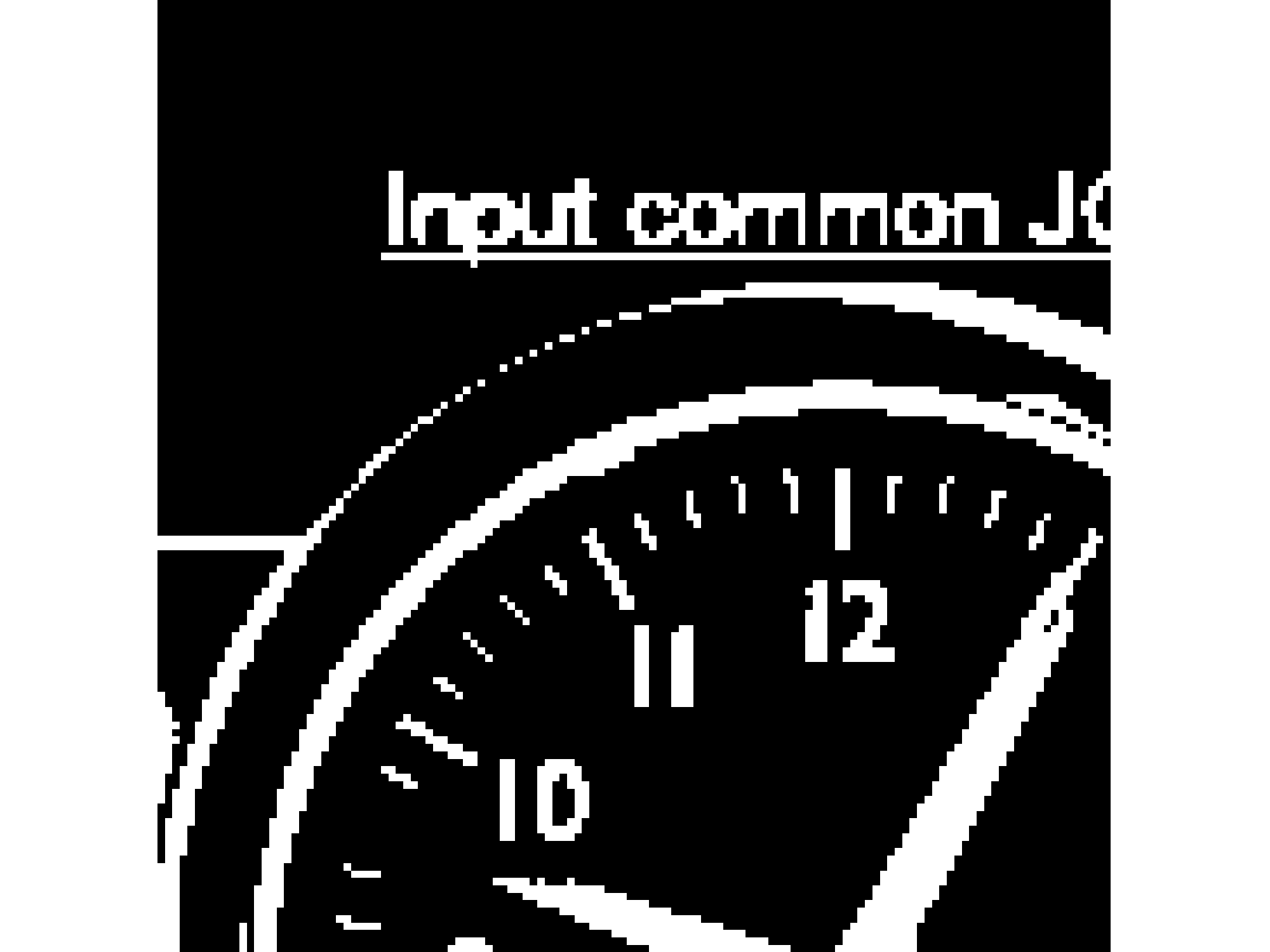}
                 \vspace{-0.45cm}
              \hspace{-4.8cm}
        \end{subfigure}
        \caption{Segmentation result for test images. The images in the first and second rows denote the original images and ground truth foregrounds. The images in the third, forth and the fifth rows denote the foreground map by shape primitive extraction and coding, hierarchical clustering in DjVu and the proposed algorithm respectively.}
\end{figure*}

%To have a numerical comparison of the proposed algorithm with DjVu and SPEC we have tested these algorithms on a dataset of 332 different image blocks of size $64\times 64$ which are extracted from
To provide a numerical comparison between the proposed scheme and previous approaches, we have calculated the average precision and recall achieved by different segmentation algorithms over this dataset. The average precision and recall by different algorithms are given in Table 1. As it can be seen, the proposed scheme achieves a much higher precision and recall than other algorithms.

\begin{table} [h]
\centering
  \caption{Accuracy comparison of different algorithms}
  \centering
\begin{tabular}{|m{1.7cm}|m{1.1cm}|m{2cm}|m{2cm}|}
\hline
Performance Criteria  &  \ SPEC & Clustering in DjVu & The proposed algorithm\\
\hline
\ \ Precision & \ 0.5038 & \ \ \ \ \ \  0.6491 & \ \ \ \ \ \ 0.9147 \\
\hline
\ \ Recall & \ 0.6458 & \ \ \ \ \ \ 0.6909 & \ \ \ \ \ \ 0.8773 \\
\hline
\end{tabular}
\label{TblComp}
\end{table}

The results for 5 test images (each consisting of multiple 64x64 blocks) are shown in Fig. 1. 
It can be seen that in all cases the proposed algorithm gives superior performance over DjVu and SPEC. 
Note that our dataset mainly consists of challenging images where the background and foreground have overlapping color ranges. For simpler cases where the background has a narrow color range that is quite different from the foreground, both DjVu and the proposed method will work well. On the other hand,  SPEC  does not work well when the background is fairly homogeneous within a block and the foreground text/lines have varying colors.

\section{Conclusion}
%\vspace{-0.5cm}
This paper proposed an algorithm for segmentation of screen content images into a foreground layer consisting of mainly text and lines and a background layer consisting of smoothly varying regions.
We developed a least absolute deviation approach to fit a smooth model into image blocks. A pixel is considered background if it can be represented accurately by the smooth model; otherwise it will be considered as foreground. 
Instead of applying this algorithm to every block, which is computationally demanding, we first check whether a block can be segmented using simpler methods. This helps to reduce the computation complexity. 
This algorithm has been tested on several test images and compared with two well-known algorithms for foreground segmentation, SPEC and hierarchical clustering in DjVu, and it shows significantly better performance for blocks where the background and foreground pixels have overlapping intensities. 
Note that the proposed algorithm is not limited to screen content image segmentation. It has other applications such as text extraction in images and principal line extraction in palmprint images.

%We like to point out that the proposed algorithm can also be applied to the cases where the background is not smooth. As long as the foreground shows a distinct characteristic from the background, we can find a set of basis functions to represent the background part, then it is possible to separate them using the proposed approach.

%\clearpage
%\section{REFERENCES}
%\label{sec:ref}

%\clearpage
%\bibliographystyle{IEEEbib}
%\bibliography{refs}
%\bibliography{sparse_ref}

\end{document}